\documentclass[acmsmall]{acmart}
\AtBeginDocument{%
  }

\setcopyright{acmlicensed}
\copyrightyear{2024}
\acmYear{2024}
\acmDOI{XXXXXXX.XXXXXXX}

\usepackage{overpic,picture}
\usepackage{placeins}
\usepackage{enumitem}
\usepackage{makecell}
\usepackage{forest}
\usepackage{tikz}
\usepackage{titlesec}
\usepackage{booktabs}
\usepackage{wrapfig}

\begin{document}

\title{The Pursuit of Fairness in Artificial Intelligence Models: A Survey}

\author{Tahsin Alamgir Kheya}
\email{s224091662@deakin.edu.au}
\affiliation{%
  \institution{Deakin University}
  \city{Waurn Ponds}
  \state{Victoria}
  \country{Australia}
  \postcode{3216}
}

\author{Mohamed Reda Bouadjenek}
\email{reda.bouadjenek@deakin.edu.au}
\affiliation{%
 \institution{Deakin University}
  \city{Waurn Ponds}
  \state{Victoria}
  \country{Australia}}

\author{Sunil Aryal}
\email{sunil.aryal@deakin.edu.au}
\affiliation{%
 \institution{Deakin University}
  \city{Waurn Ponds}
  \state{Victoria}
  \country{Australia}
}
  
\renewcommand{\shortauthors}{Kheya, et al.}

  \begin{abstract}

Artificial Intelligence (AI) models are now being utilized in all facets of our lives such as healthcare, education and employment. Since they are used in numerous sensitive environments and make decisions that can be life altering, potential biased outcomes are a pressing matter. Developers should ensure that such models don't manifest any unexpected discriminatory practices like partiality for certain genders, ethnicities or disabled people. With the ubiquitous dissemination of AI systems, researchers and practitioners are becoming more aware of unfair models and are bound to mitigate bias in them. Significant research has been conducted in addressing such issues to ensure models don't intentionally or unintentionally perpetuate bias. This survey offers a synopsis of the different ways researchers have promoted fairness in AI systems. We explore the different definitions of fairness existing in the current literature. We create a comprehensive taxonomy by categorizing different types of bias and investigate cases of biased AI in different application domains. A thorough study is conducted of the approaches and techniques employed by researchers to mitigate bias in AI models. Moreover, we also delve into the impact of biased models on user experience and the ethical considerations to contemplate when developing and deploying such models. We hope this survey helps researchers and practitioners understand the intricate details of fairness and bias in AI systems. By sharing this thorough survey, we aim to promote additional discourse in the domain of equitable and responsible AI.
\end{abstract}





\maketitle

\section{Introduction}

\begin{figure}[t]
    \begin{overpic}[
      width=\linewidth]{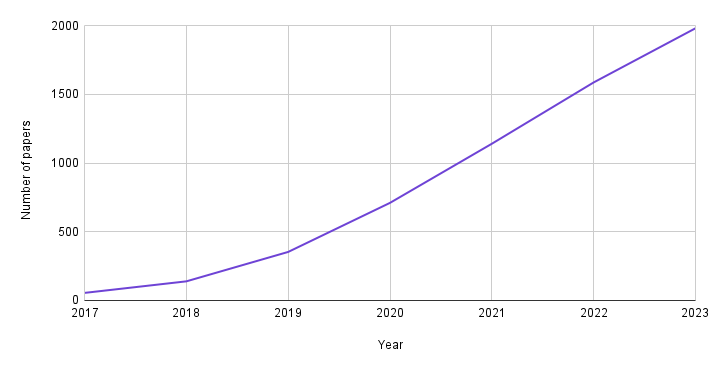}
  
    \end{overpic}
  \caption{Number of papers published in this topic over the years. Data acquisition process to plot this graph is provided in Section \ref{psp}.}
        \label{fig:ml-comp}
\end{figure}

The use of automated systems has rapidly advanced across various domains, influencing everything from hiring employees to recommendation systems. AI systems are embedded in our day-to-day activities and influence our lives greatly, especially when used to make life-altering decisions. These models have great potential, as they can integrate tons of data and perform very complex computations more effectively and faster than humans. Amid AI's potential, however, concerns arise regarding the fairness and bias in these systems. Since these systems are being used in sectors like healthcare, finance, and criminal justice to make prominent decisions for individuals, ensuring fairness in these models is critical. 
\hfill\\
In recent years a number of cases of AI bias has been exposed and the major effect it has on individuals and communities is inevitable. For instance, in the USA an algorithm used to find the recidivism score for sentencing was found to be biased towards Black defendants \cite{mattu_machine_nodate}. Google Bard was seen to depict gender stereotypes by stating boys want to achieve goals and make a difference in life while girls want love and affection\cite{fowler_perspective_2023}. These are just two examples, but numerous concerns like these have led to growing interest in developing and deploying fair AI models. Fig \ref{fig:ml-comp} shows the number of papers published in this field for the last seven years. Over the years the volume of papers published in this domain has steadily increased. By 2021, the number of papers surged over 1000. The steady increase resulted in numbers close to a whopping 2000 papers published last year.
The graph underscores the significance of these topics in the research community over the years. 
\hfill\\
When evaluating models' fairness more than one definition of fairness has been used. This survey explores all the different fairness criteria discussed in the literature. Several researchers have been working extensively to address fairness issues in automated models. With the broad domain of fair AI, researchers have put forward multiple strategies to address and mitigate bias in them. It is also important to realize that certain strategies only work on certain types of bias. This paper thoroughly describes the different types of biases and all the common approaches used to mitigate these biases. Moreover, this survey covers details on the causes of unfairness, different cases of biases within different sectors including but not limited to healthcare, education and finance. Working towards making AI models fair can also enhance user experience. In this paper, we discuss the impact of biased models on users and ethical guidelines that should be followed to ensure users' trust. At the end of the paper, we mention the challenges and limitations of the current literature. Overall the aim of the paper is to shed light on the existing work done on bias and fairness in the context of AI models. We hope this paper will provide researchers and practitioners with enriched perspectives in this field and encourage them to decide on their research direction and develop innovative ideas to mitigate unintended consequences.

\FloatBarrier

\section{Related Survey}
In recent years bias and fairness in the context of machine learning has been a hot topic. Numerous surveys have been written in this domain. These recent surveys \cite{mehrabi_survey_2021, ferrara2023fairness,pessach, simon23, tangg23,ntoutsi_bias_2020,app131810258} are thorough and provide insight into the cause of bias, mitigating them and promoting fairness in a more general context. There are several papers that explore fairness in a specific domain which include recommender systems \cite{deldjoo_fairness_2023,chen2023}, vision language models \cite{lee_survey_2023,parraga23}, healthcare \cite{ueda_fairness_2024,fletcher_addressing_2021,adela23}, finance \cite{pavon_perez_bias_2022} and NLP \cite{blodgett_language_2020,bansal2022survey,app11073184}.  Our survey incorporates recent research which includes more updated definitions of bias and fairness, mitigation strategies and causes of unfairness in real life cases. Moreover almost none of the previous surveys delves into details about how user experience was impacted because of bias in ML models and how this phenomenon has affected different sectors like education, recruitment etc. This paper thoroughly explores these neglected sub-topics.
Table \ref{tab:comparison}, showcases the subjects that were covered by other papers (akin to this paper) and some that were overlooked. The table is divided into 2 sections. The first few rows describe papers that capture the broad domain of fairness and bias in AI, and the second part describes surveys that cover specific domains like healthcare, NLP, and more.
\begin{table}[h] 
  \caption{A comparison of topics covered and overlooked by the recent surveys resembling this paper}
  \label{tab:comparison}
  \resizebox{\textwidth}{!}{ \begin{tabular}{cccccccccc l}
    \toprule
   Year & Paper & Fairness  & Types of   & Causes of & Cases of & Biased Models within  & Mitigation   & The Impact Bias has on & Ethical  \\
   &  &  Definition &  Bias  &  Unfairness &  Bias & Different Sectors  &  Strategies  & User Experience &  Considerations \\

    \midrule
     &This Paper & \checkmark & \checkmark & \checkmark & \checkmark & \checkmark & \checkmark & \checkmark & \checkmark\\
       2020 &\cite{ntoutsi_bias_2020} & \checkmark & x & \checkmark & \checkmark & x & \checkmark & x & x  \\
     2021 &\cite{mehrabi_survey_2021} & \checkmark & \checkmark & \checkmark & \checkmark & x & \checkmark & x & x  \\
        2022 &\cite{pessach} & \checkmark & x & \checkmark & \checkmark & x & \checkmark & x & x  \\
        
   2023 &\cite{simon23} & \checkmark & x & \checkmark & x & x & \checkmark & x & x  \\
   
    2023 &\cite{tangg23} & \checkmark & x & \checkmark  & x & x & \checkmark & x & x  \\
 
   2023  &\cite{ferrara2023fairness} & \checkmark & \checkmark & \checkmark & \checkmark & x & \checkmark & \checkmark & x  \\
   2023  &\cite{app131810258} & \checkmark & \checkmark & \checkmark & \checkmark & x & \checkmark & x & x  \\
    2023  &\cite{digital4010001} & x & \checkmark & \checkmark & \checkmark & x & \checkmark & x & x  \\

      2023  &\cite{kaur22} & x & \checkmark & x & x & x & \checkmark & x & \checkmark  \\
        \midrule
        \multicolumn{10}{c}{\textbf{Specialized surveys}} \\
        \midrule
    2020 &\cite{blodgett_language_2020} & x & x & \checkmark & \checkmark & x & \checkmark & x & x  \\
    2021 &\cite{fletcher_addressing_2021} & \checkmark & \checkmark & \checkmark & \checkmark & x & \checkmark & x & x  \\
    2021 &\cite{nima22} & x & \checkmark & \checkmark & \checkmark & x & \checkmark & x & x  \\
    2021 &\cite{kaur2021} & x & \checkmark & \checkmark & \checkmark & x & \checkmark & x & \checkmark  \\
    2021 &\cite{perr21} & \checkmark & x & \checkmark & \checkmark & x & \checkmark & x & x  \\
    2021 &\cite{app11073184} & x & \checkmark & \checkmark & \checkmark & x & \checkmark & x & x  \\
    2022 &\cite{lee_survey_2023} & \checkmark & x & \checkmark & \checkmark & x & \checkmark & x & x  \\
    2022 &\cite{bansal2022survey} & \checkmark & \checkmark & x & x & x & \checkmark & x & x  \\
  
    2023 &\cite{adela23} & x & \checkmark & \checkmark & \checkmark & x & \checkmark & x & x  \\
    2023 &\cite{correa_systematic_2022} & x & \checkmark & \checkmark & \checkmark & x & \checkmark & x & x  \\
    2023 &\cite{deldjoo_fairness_2023} & \checkmark & \checkmark & \checkmark & \checkmark & x & \checkmark & x & x  \\
    2023 &\cite{parraga23} & \checkmark & x & \checkmark & \checkmark & x & \checkmark & x & x  \\
    2024 &\cite{ueda_fairness_2024} & x & \checkmark & \checkmark & \checkmark & x & \checkmark & x & \checkmark  \\
 
  \bottomrule
\end{tabular}}
 
\end{table}

\section{ Conceptualizing Fairness and Bias in ML}
In machine learning, fairness and bias are intertwined concepts that follow the same aspect: how models' predictions can favor a particular group of people. Bias in the model will lead to unfairness; to get a fair model, we must mitigate bias. In simpler terms, bias is the issue, and fairness is the solution. Although fairness is stated to be the solution, it is important to note that achieving absolute fairness is a very challenging task, primarily because of the different fairness criteria. So, till now, there has been no single solution that mitigate all types of bias and makes a model absolutely fair.

\tikzset{
    my node/.style={
        thick,
        minimum height=1cm,
        minimum width=1cm,
        text width=25ex,
        text height=0ex,
        text depth=0ex,
        font=\sffamily,
    }
}

   \begin{figure}[ht]
    \begin{forest}
    for tree={
        my node,
        parent anchor= east,
        grow' = east,
        draw=violet, 
        child anchor = west,
        l sep = 15pt,
        rounded corners=5pt,
        edge path={
            \noexpand\path[\forestoption{edge}]
            (!u.parent anchor) -- +(5pt,0) |- (.child anchor)\forestoption{edge label} [violet];
        },
    },
    [Fairness in Machine Learning ,text depth=1ex
        [Group Fairness {\textcolor{violet}{Sec: \ref{grp-f}}}
            [Demographic parity {\textcolor{violet}{Sec: \ref{demo-p} }}, text depth=1ex]
            [Conditional Statistical Parity {\textcolor{violet}{Sec: \ref{csp}}}, text depth=1ex]
        ] 
        [Individual Fairness {\textcolor{violet}{Sec: \ref{ind-f}}}
            [Fairness through Unawareness {\textcolor{violet}{Sec: \ref{ftu}}}, text depth=1ex]
            [Fairness through Awareness {\textcolor{violet}{Sec: \ref{fta}}}, text depth=1ex]
        ]
        [Separation Metrics {\textcolor{violet}{Sec: \ref{sep-m}}}
            [Predictive Equality {\textcolor{violet}{Sec: \ref{pred-e}}} , text depth=0.5ex]
            [Equal Opportunity {\textcolor{violet}{Sec: \ref{eq-op}}}, text depth=0.5ex]
            [ Balance for the Negative Class {\textcolor{violet}{Sec: \ref{bnc}}}, text depth=1ex]
            [ Balance for the Positive Class {\textcolor{violet}{Sec:  \ref{bnp}}}, text depth=1ex]
            [Equalized Odds {\textcolor{violet}{Sec:  \ref{equalized-odds}}}, text depth=0.5ex]
        ] 
        [Intersectional Fairness {\textcolor{violet}{Sec:  \ref{int-f}}},text depth=1ex]
        [Treatment Equality {\textcolor{violet}{Sec: \ref{treat-e}}}]
        [Sufficiency Metrics {\textcolor{violet}{Sec: \ref{suff-m}}}
            [Equal Calibration {\textcolor{violet}{Sec: \ref{eq-cal}}} , text depth=0.5ex]
            [Predictive Parity {\textcolor{violet}{Sec: \ref{pred-p}}} , text depth=0.3ex]
            [Conditional Use Accuracy Equality {\textcolor{violet}{Sec: \ref{cuae}}}, text depth=1.7ex]
        ]
        [Causal-based Fairness {\textcolor{violet}{Sec:  \ref{cas-f}}}, text depth=1ex
            [Counterfactual Fairness {\textcolor{violet}{Sec: \ref{count-f}}}, text depth=1ex]
            [Unresolved Discrimination {\textcolor{violet}{Sec: \ref{unres-disc}}}, text depth=1ex]
        ]
    ]
    \end{forest}
      
  \caption{Proposed taxonomy of fairness in the machine learning context}
        \label{fig:ml-fairness}
\end{figure}
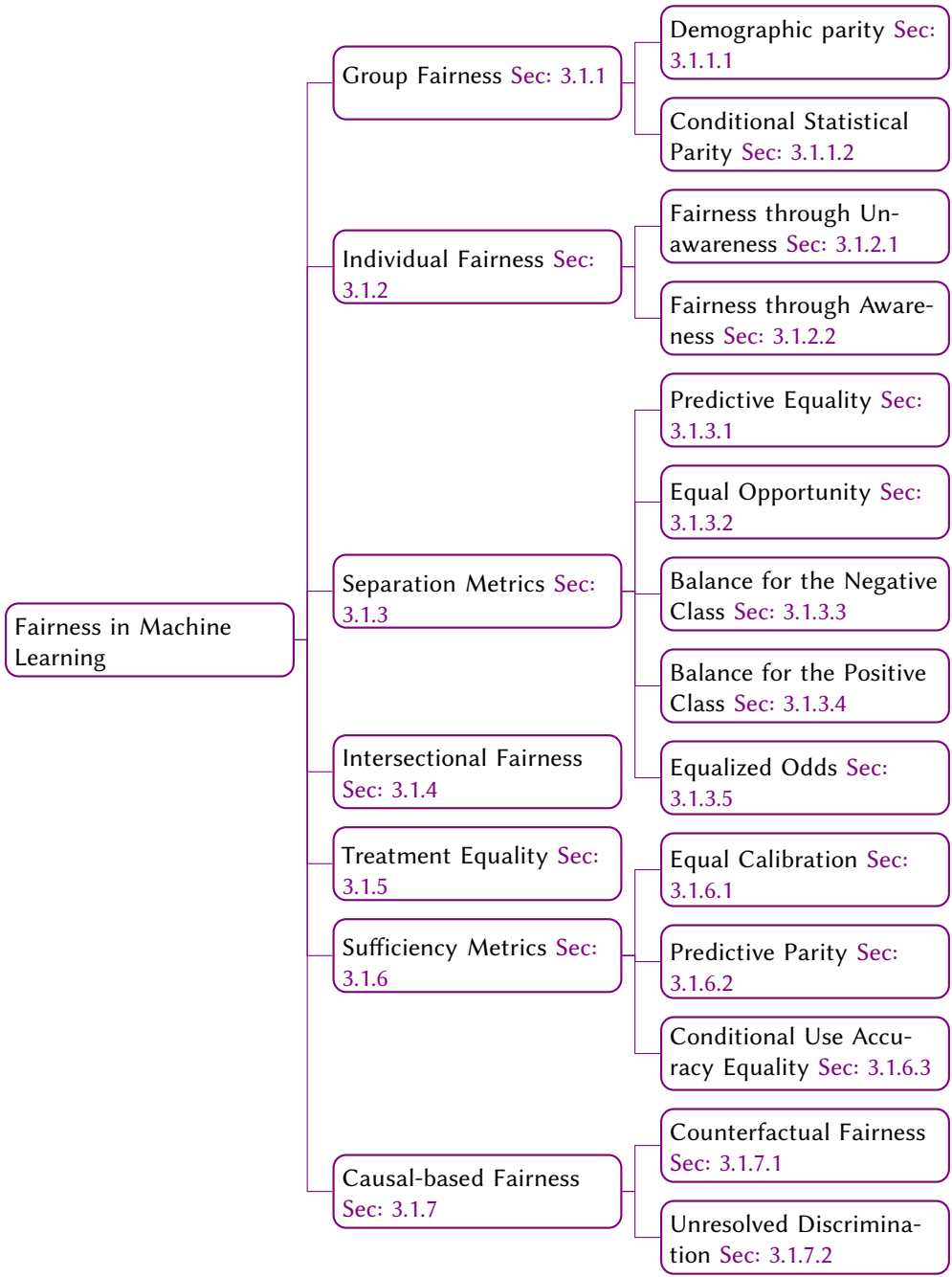
\FloatBarrier

\subsection{Fairness in Machine Learning}
 In simple terms, fairness in machine learning refers to models treating individuals and groups in an ethical manner while making predictions. In recent years, extensive research has been conducted on producing fair models based on different definitions of fairness. Figure \ref{fig:ml-fairness} presents a taxonomy of fairness definitions utilized in previous research. It is compiled based on the concept of fairness discussed in existing literature such as \cite{fino21, saxena19,calegari23,gohar2023}.

   
\subsubsection{Group fairness}\label{grp-f}
 \hfill\\
 Group Fairness, in the context of machine learning, describes the phenomena of the model treating different groups equally. For example, the ML model should treat people of different genders impartially. The concept of group fairness can be subdivided into various definitions.

\paragraph{ Demographic Parity}\label{demo-p}
\hfil\\
Demographic parity ensures that the predicted outcome \(\hat{Y}\) of an ML model is independent of sensitive attribute $S$ \cite{grari_fairness-aware_2020}. For example, a model's prediction of a potential candidate to be hired for a position should not depend on the candidate's gender. For binary classification with a binary sensitive attribute $S$, demographic parity can be formalized as \cite{hardt2016}: \[P(\hat{Y}=1|S=0)=P(\hat{Y}=1|S=1)\].

\paragraph{Conditional Statistical Parity}\label{csp}
 \hfill\\
 Conditional statistical parity, also known as conditional demographic parity, requires the outcome for different sensitive groups to be the same, even after adding extra features \cite{Corbett17}. For binary classification with binary sensitive attribute $S$, and $F$ which is another feature (where $f$ is the value of this feature) conditional statistical parity can be formalized as:
 \[ P(\hat{Y}=1 | S=0, F=f)=P(\hat{Y}=1 | S=1, F=f)\]
 For example, for a model to decide if a student should be admitted to a school, the additional features considered could be GPA and admission test scores. Conditional statistical parity is satisfied as long as a similar number of male and female students are admitted for any combination of academic performance.

    \subsubsection{Individual fairness}\label{ind-f}
 \hfill\\
    Individual fairness focuses on ensuring that similar individuals receive similar predictions from a ML model, regardless of their membership in protected groups. In other words, this definition ensures individuals are characterized by their individual traits and not by any group stereotypes. Fairness through awareness and fairness through unawareness both fall under the umbrella of individual fairness because they are related to how individuals are treated by the ML model on a one-to-one basis.
    
    \paragraph{Fairness through Unawareness}\label{ftu}
     \hfill\\
    A model is said to follow this definition of fairness as long as it doesn't use any sensitive attributes when making decisions \cite{kusner_counterfactual_2018}.  The equation \[ Y: X \to \hat{Y}\] represents the function of the model (which takes input X and predicts outcome $\hat{Y}$) \cite{kusner_counterfactual_2018}. The main point here is that this mapping should exclude any sensitive attribute $S$. Although this definition is very simple, other features used to train the model can contain discriminatory information \cite{kusner_counterfactual_2018,grari_fairness-aware_2020,calegari23}. This can lead the model to infer attributes like gender, ethnic background, etc., from these features, leading to unfair predictions.
    
    \paragraph{Fairness through Awareness}\label{fta}
    \hfill\\
    The concept of fairness through awareness is a bit more sophisticated. This approach explicitly considers the sensitive attributes when the model is trained. Here $k$ is a metric to compute the similarity of candidates and $M$ is a model/function to predict their selection probabilities. This fairness criterion will hold if for any two candidates $a$ and $b$, the difference in the distribution assigned to them (denoted by D(Ma,Mb)) is less than or equal to their similarity (denoted by k(a,b), i.e., \[D(Ma, Mb\leq k(a,b)\] 
    \subsubsection{Separation Metrics}\label{sep-m}
     \hfill\\
     Separation metrics are a set of statistical criteria that try to enforce fairness by evaluating the model. The model is assessed to check the extent to which it separates the outcomes for different classes. There are several separation metrics, which are discussed in the subsections below.
 \paragraph{Predictive Equality}\label{pred-e}
 \hfill\\
 A model is said to satisfy this definition of fairness if the FPR (False Positive Rate) is equal for both protected group and unprotected group \cite{verma2018, Corbett17}. For example, the probability of an individual with an actual bad credit score being incorrectly given a good credit score should be equal in different subgroups of a sensitive attribute \cite{verma2018}. With $\hat{Y}$ representing the prediction, $S \in \{0,1\}$ a sensitive attribute and $Y$ the actual outcome, this definition can be formalized as:
 \[
P(\hat{Y} = 1 | S = 0, Y = 0) = P(\hat{Y} = 1 | S = 1, Y = 0)
\]
\paragraph{ Equal Opportunity}\label{eq-op}
 \hfill\\
This concept of fairness ensures that individuals from different groups have an equal chance of receiving a positive outcome. In simple terms, this concept ensures individuals who obtain the "advantaged" outcome (e.g. successful candidate to be hired) have an equal chance of getting this prediction, regardless of any protected attribute \cite{hardt2016}. This definition can be formalized for a binary classifier as \cite{hardt2016}:
\[
P(\hat{Y} = 1 | S = 0, Y = 1) = P(\hat{Y} = 1 | S = 1, Y = 1)
\]
It is quite similar to predictive equality but the focus for this definition is the true positive rate balance, whereas for predictive equality it is the false positive rate balance.
\paragraph{ Balance for the Negative Class}\label{bnc}
\hfill\\
This concept of fairness ensures that the predicted scores assigned by the model to individuals belonging to the negative class are the same for both protected and unprotected groups \cite{verma2018}. With an average predicted probability score of Z, binary sensitive attribute S $\in \{0,1\}$ and predicted outcome  $\hat{Y} $  this definition can be formalized as:
 \[
P(Z | \hat{Y} = 0 , S=1) = P(Z | \hat{Y} = 0 , S = 0)
\]
  For example, for this definition to be satisfied, a model used to hire teachers should provide the same scores of not being hired for the candidates, if deemed not suitable, regardless of the race of the individuals.

\paragraph{ Balance for the Positive Class}\label{bnp}
 \hfill\\
 This concept tries to ensure that the same predicted scores are assigned by the model to individuals belonging to the positive class for both protected and unprotected groups \cite{verma2018}. With a predicted probability score Z, binary sensitive attribute S $\in \{0,1\}$ and predicted outcome  $\hat{Y}$  this definition can be formalized as:
 \[
P(Z | \hat{Y} = 1 , S=1) = P(Z | \hat{Y} = 1 , S = 0)
\]
Let's consider a similar example to the last definition (balance for negative class). For this notion of fairness to be met, a model used to hire teachers should provide the same scores of being hired for the candidates, if deemed suitable regardless of the race of the candidate.
 
\paragraph{Equalized Odds
}\label{equalized-odds}
 \hfill\\
This concept of fairness holds if the predictor $\hat{Y}$ and sensitive attribute $S$ are independent given $Y$ \cite{hardt2016}. $\hat{Y}$ is allowed to depend on $S$, when genuinely relevant to the outcome. This metric also ensures that the true positive rates and false positive rates are equal for different groups of individuals \cite{verma2018}. This promotes fairness whilst still allowing the model to leverage useful insights by not omitting sensitive attributes (only when appropriate). According to \cite{hardt2016}, this definition can be formalized as:
\[
P(\hat{Y} = 1 | S = 0, Y = y) = P(\hat{Y} = 1 | S = 1, Y= y), \quad y \in \{0,1\}
\]
Let's consider a model that suggests restaurants for you to eat. The sensitive attribute $S$ in this case is the religion of users. When training, the model is allowed to consider $S$, but only if it is relevant to the true outcome. It might learn that individuals following religion $R$ tend to like fish and other sea foods. However, it cannot simply suggest just pescatarian options to individuals who follow religion $R$. This is to make sure that the model bases its predictions on appropriate features like dietary restrictions, price range, location, etc.

\subsubsection{ Intersectional Fairness}\label{int-f}
\hfill\\
Intersectional fairness acknowledges that individuals can encounter some form of discrimination because of their overlapping identities. This concept goes beyond traditional fairness notions and considers more than individual characteristics. According to \cite{gohar2023}, intersectional identities can intensify unfairness that's not even present in constituent groups (like Black woman vs Black vs woman). Readers interested in the realm of intersectional fairness are directed to review \cite{gohar2023,jamesr18}. 

    \subsubsection{Treatment Equality}\label{treat-e}
 \hfill\\
    This concept of fairness aims to achieve equal proportions of false negatives to false positives for both unprotected and protected groups \cite{berk2017fairness}. This definition can be formalized as:
     \[
\frac{FN_1}{FP_1}= \frac{FN_2}{FP_2}
\] where subscripts 1 and 2 represent protected and unprotected groups, respectively.

    \subsubsection{Sufficiency Metrics}\label{suff-m}
 \hfill\\
 These metrics ensure that a model is equally calibrated to make fair decisions for different sensitive groups, like ethnicity, religion, age etc. This idea can be subdivided into three definitions. These are discussed in the next sections.
 
\paragraph{Equal Calibration}\label{eq-cal}
 \hfill\\
 This concept of fairness ensures that for a given probability score $Z$, people in both protected and unprotected categories should possess an equal probability of being in the positive class \cite{chouldechova2016fair}. This definition is formalized by \cite{chouldechova2016fair} as:
  \[
P(\hat{Y}=1 | Z=z, S=0) = P(\hat{Y}=1 | Z=z, S=1)
\]
where, $\hat{Y}$ is the predicted outcome, $Z$ is the score, and $S$ is the sensitive attribute. This definition is pretty similar to predictive parity (refer to \ref{pred-p}) since they ensure accuracy for both groups but equal calibration applies beyond binary scores \cite{chouldechova2016fair}. For instance, this definition holds if, for any given score $z$, a model used to hire candidates, predicts the same chance of actually getting hired for different genders.
 
 \paragraph{ Predictive Parity}\label{pred-p}
 \hfill\\
 This concept of fairness holds if the PPVs (Positive Predictive Value) for protected and unprotected groups are equal \cite{verma2018,chouldechova2016fair}. This means an individual who was predicted to get a positive outcome should actually get a positive outcome. As stated by the authors in \cite{verma2018}, this definition can be formalized as 
  \[
P(Y=1 | \hat{Y}=1, S=0) = P(Y=1 | \hat{Y}=1, S = 1)
\]
where $\hat{Y}$ is the predicted outcome, $Y$ is the true outcome and $S$ is the sensitive attribute. Let's think about a model which is used to determine if a person will repay a loan. For this definition to be satisfied, the probability of being in a positive group (pay back loan) should be the same as their likelihood of actually paying back the loan.

 \paragraph{Conditional Use Accuracy Equality}\label{cuae}
 \hfill\\
 This definition of fairness ensures that PPVs (Positive Predicted values) and NPVs (Negative Predicted Values) are the same regardless of what sensitive group an individual belongs to \cite{verma2018,chouldechova2016fair}. Here, NPV describes the probability of an individual being predicted with a negative outcome actually being in the negative class. A formal definition of NPV is given as: \[NPV= \frac{True Negatives}{True Negatives+False Negatives}\] Here, PPV describes the probability of an individual being predicted with a positive outcome actually being in the positive class. 
 A formal definition of PPV is given as: \[PPV= \frac{True Positives}{True Positives+False Positives}\]
 Let's consider a model that is used to decide if a person will repay a loan taken. For this definition to hold:
 \hfill\\
 a. Throughout all sensitive groups who are predicted to be in the negative group (doesn't pay loan), the actual rate of not paying back the loan is the same.
 \hfill\\
b. Throughout all sensitive groups who are predicted to be in the positive group (pay back loan), the actual rate of paying the loan is the same.
\hfill\\
The authors in \cite{verma2018} formalizes this definition as :
 \[
(P(Y = 1 | \hat{Y}=1, S=0) = P(Y =1 | \hat{Y}=1, S=1))  \land  (P(Y = 0 | \hat{Y}=0, S=0) = P(Y=0 | \hat{Y}=0, S=1))
\]
where $\hat{Y}$ is the predicted outcome, $Y$ is the actual outcome and $S$ is the sensitive attribute.

    \subsubsection{Causal-based Fairness}\label{cas-f}
 \hfill\\
    This concept involves using additional knowledge, like insights from experts, to identify the causal structure of a particular case \cite{calegari23}. For example, exploring hypothetical situations and asking questions like "what would happen if an individual had a different race" \cite{calegari23}. This fairness concept is broken down into two further sub-categories discussed next.
   
 \paragraph{ Counterfactual Fairness}\label{count-f}
 \hfill\\
For a model to be counter-factually fair, it needs to have the same predictions for individuals having the same relevant features even if the protected attributes are different. As stated by \cite{kusner_counterfactual_2018}, this definition can be formalized as:
\[ P(\hat{Y}_{S\leftarrow s}(U) = y | X = x, S = s) = P(\hat{Y}_{S\leftarrow s'}(U) = y  | X = x, S = s)
\]
where $S$ is the protected attribute, $U$ is the set of latent background variables, $X$ is the remaining attributes (which is under context $X=x$ and $S=s$). Here $\hat{Y}_{S\leftarrow s}(U)$ represents the counterfactual variable $\hat{Y}$, when S is set to s by an external intervention \cite{marco22}.

 \paragraph{Unresolved Discrimination}\label{unres-disc}
 \hfill\\
 This kind of discrimination can arise when a sensitive attribute unfairly impacts the predicted outcome. In a causal graph, variable V can exhibit unresolved discrimination if a directed path exists from S (a sensitive attribute) to V that is not blocked by a resolving variable \cite{kilbertus2018avoiding}. 
  \begin{wrapfigure}{r}{0.3\textwidth}
   \begin{overpic}[
      width=0.6\linewidth]{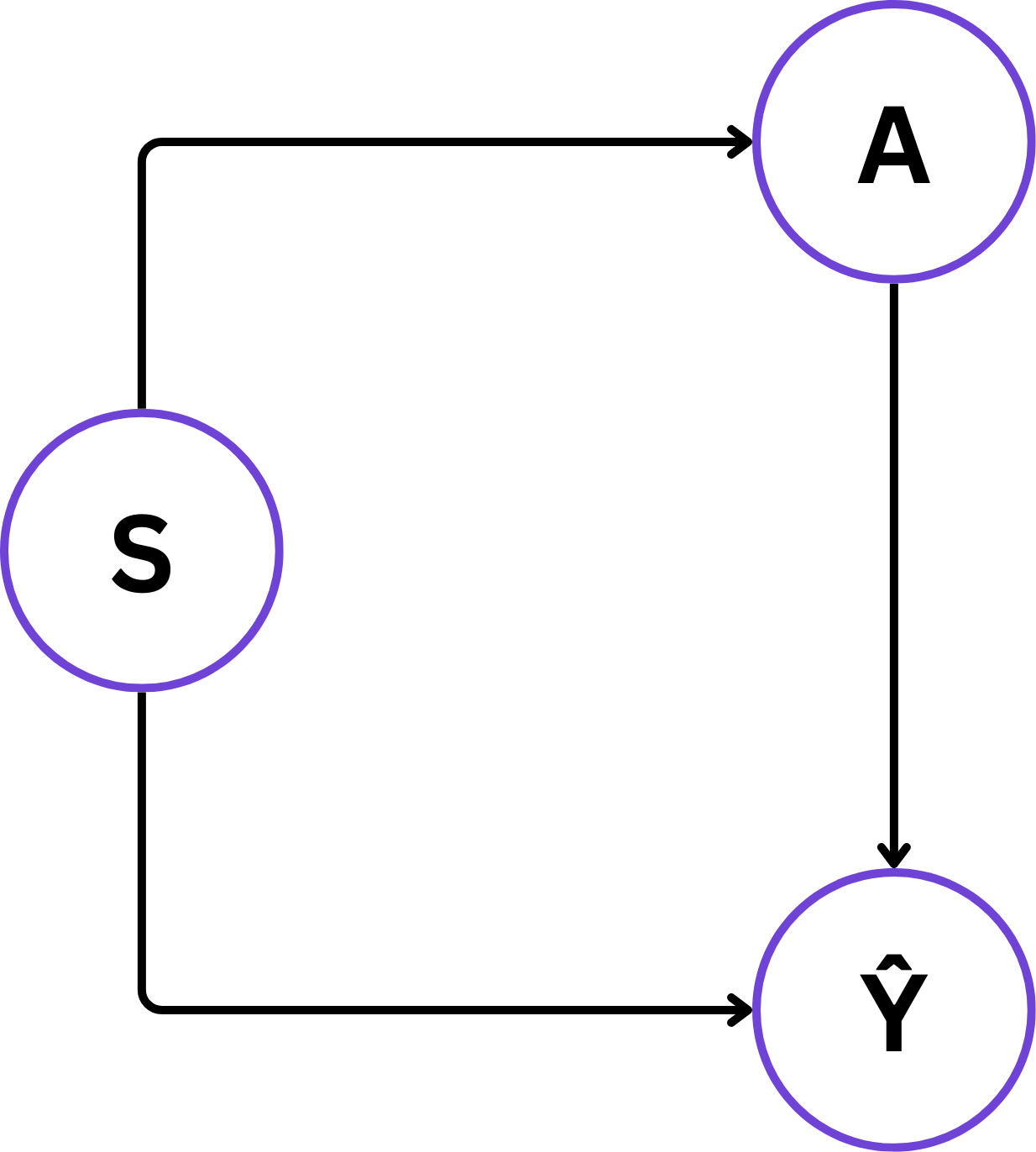}
     
    \end{overpic}
  
    \caption{Graph that exhibits unresolved discrimination. }
    \label{fig:causal graph example}
\end{wrapfigure}
 For this to be true V itself should be a non-resolving variable. A resolving variable can be described as a variable that intervenes between a sensitive attribute and the predicted outcome with the intentions to justify any observed discrimination.

Figure \ref{fig:causal graph example} shows a causal graph where race (S) is directly impacting the decision of the housing application (A), that cannot be attributed by any resolving variable. Housing choice isn't a resolving variable, and since it is directly impacted by the sensitive attribute S it is causing an unresolved discrimination.

\subsection{Bias in Machine learning}
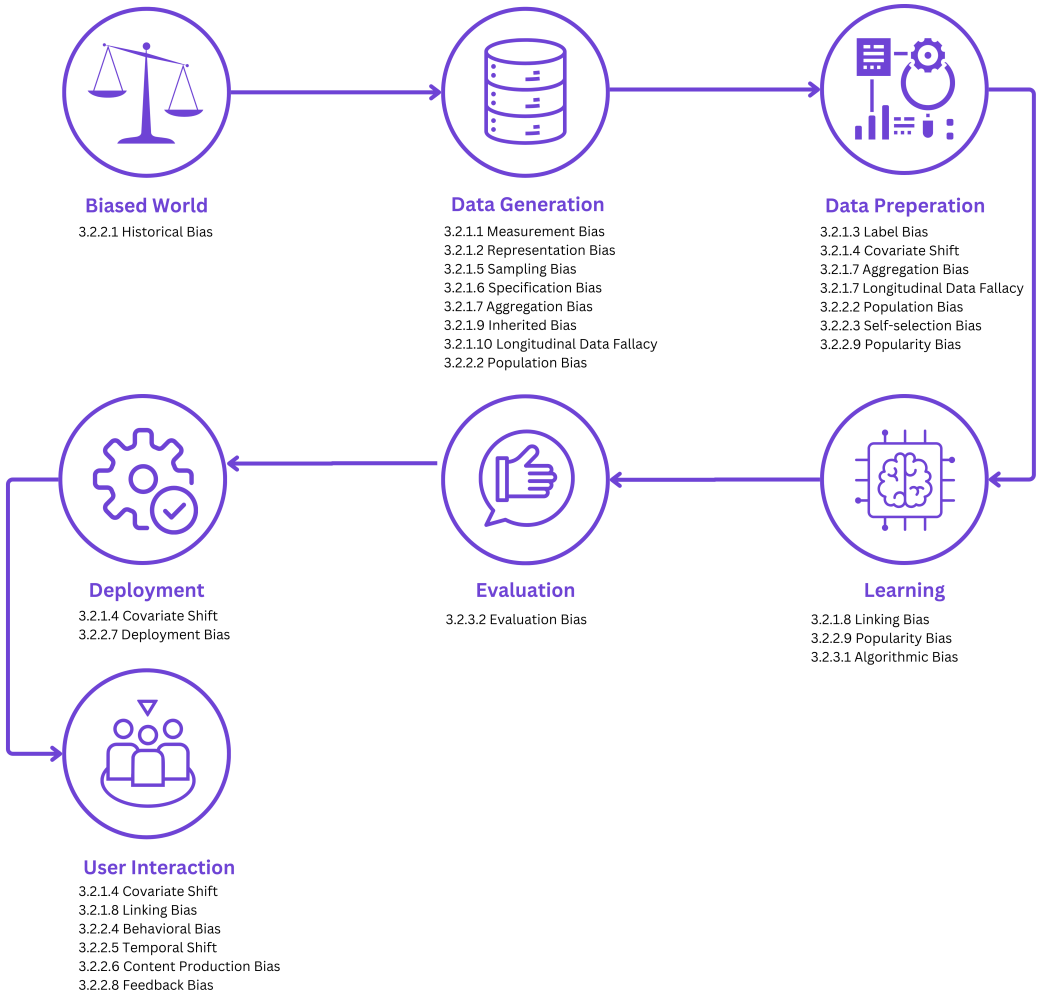
\begin{figure}[ht]
    \begin{overpic}[
      width=\linewidth]{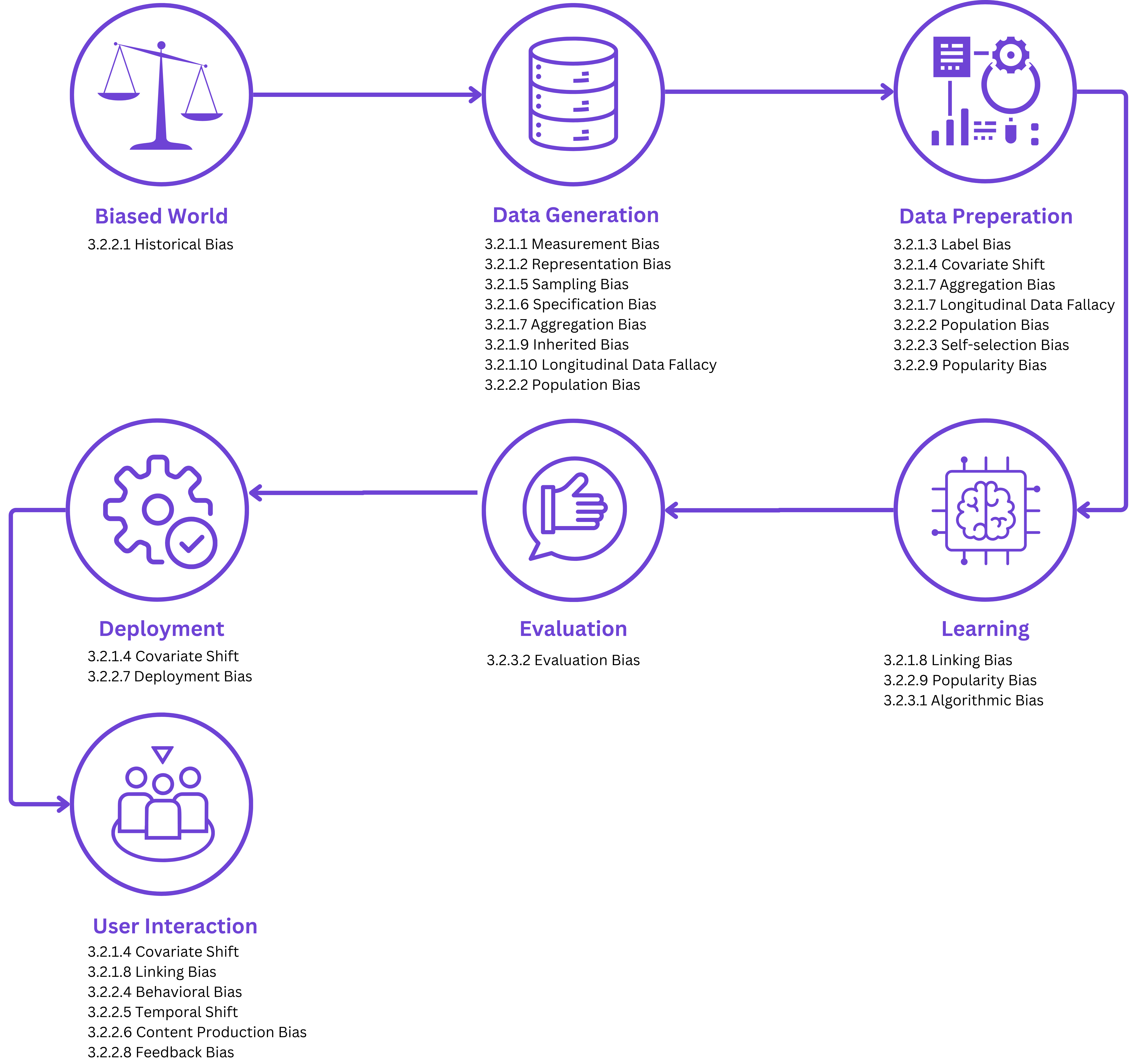}
      \put(9,70){\hyperref[historical-bias]{\makebox(10,2)[lb]{}}}%
      \put(42,72){\hyperref[measurement-bias]{\makebox(13,0.6)[lb]{}}}%
      \put(42,70){\hyperref[rep-bias]{\makebox(13,0.6)[lb]{}}}%
      \put(42,68){\hyperref[sampling-bias]{\makebox(13,0.6)[lb]{}}}%
       \put(42,67){\hyperref[spec-bias]{\makebox(13,0.6)[lb]{}}}%
        \put(42,65){\hyperref[agg-bias]{\makebox(13,0.6)[lb]{}}}%
        \put(42,63){\hyperref[inherited-bias]{\makebox(13,0.6)[lb]{}}}%
        \put(42,61){\hyperref[log-bias]{\makebox(13,0.6)[lb]{}}}%
        \put(42,59){\hyperref[pop-bias]{\makebox(13,0.6)[lb]{}}}%
        
        \put(80,72){\hyperref[lab-bias]{\makebox(8,0.6)[lb]{}}}%
        \put(80,70){\hyperref[co-bias]{\makebox(13,0.6)[lb]{}}}%
        \put(80,68){\hyperref[agg-bias]{\makebox(13,0.6)[lb]{}}}%
        \put(80,66){\hyperref[log-bias]{\makebox(13,0.6)[lb]{}}}%
        \put(80,65){\hyperref[pop-bias]{\makebox(13,0.6)[lb]{}}}%
        \put(80,63){\hyperref[ss-bias]{\makebox(13,0.6)[lb]{}}}%
        \put(80,61){\hyperref[poty-bias]{\makebox(13,0.6)[lb]{}}}%

        \put(9,36){\hyperref[co-bias]{\makebox(13,0.6)[lb]{}}}%
        \put(9,34){\hyperref[dep-bias]{\makebox(13,0.6)[lb]{}}}%
        
        \put(42,36){\hyperref[eva-bias]{\makebox(13,0.6)[lb]{}}}%

        \put(80,32){\hyperref[alg-bias]{\makebox(13,0.6)[lb]{}}}%
        \put(80,34){\hyperref[poty-bias]{\makebox(13,0.6)[lb]{}}}%
        \put(80,35){\hyperref[link-bias]{\makebox(13,0.6)[lb]{}}}%
        
        \put(8,10){\hyperref[co-bias]{\makebox(13,0.6)[lb]{}}}%
        \put(8,8){\hyperref[link-bias]{\makebox(13,0.6)[lb]{}}}%
         \put(8,6){\hyperref[bb]{\makebox(13,0.6)[lb]{}}}%
          \put(8,4){\hyperref[ts]{\makebox(13,0.6)[lb]{}}}%
          \put(8,3){\hyperref[cpb]{\makebox(13,0.6)[lb]{}}}%
          \put(8,1){\hyperref[fee-bias]{\makebox(13,0.6)[lb]{}}}%

    \end{overpic}
  \caption{Proposed taxonomy of observed biases in the machine learning pipeline}
        \label{fig:ml-bias}
\end{figure}
       
The word `bias' is derived from the French word `biais', which means slope. Later, this word's meaning was expanded in English and is now used to refer to the inclination towards supporting or opposing a certain person or group in an unfair way. In Machine Learning (ML), the same concept is applicable. This section discusses different types of biases that can arise in the ML pipeline. Figure \ref{fig:ml-bias} shows the various biases that arise at every step of the ML pipeline. The list is compiled from existing research works and includes concepts mentioned in \cite{fahse_managing_2021,mehrabi_survey_2021, gu2019understanding, ThomasH, Suresh_2021}. The various types of biases are organized in 3 sections, which include data-driven, human and model bias.


    \subsubsection{Data-Driven Bias}\label{ddb}
 \hfill\\
    This section describes biases that arise in the outcomes of models that learn patterns from the training data.
    \paragraph{Measurement Bias}\label{measurement-bias}
 \hfill\\
    Measurement bias is introduced when subjective choices are made for the model design \cite{fahse_managing_2021}. This includes the act of selecting, gathering or computing features and annotations to be used in a prediction problem \cite{Suresh_2021}. One case of this kind of bias occurred when an algorithm was used to estimate grades of students who couldn't sit for proper exams during COVID-19. This algorithm used historical school performance as a proxy to predict the abilities of the current students. It introduced measurement bias that unfairly affected students' grades, particularly those with lower historical performance \cite{denes_case_2023}.
    
    \paragraph{Representation Bias}\label{rep-bias}
 \hfill\\
    Representation bias can occur during the data collection or sampling phase when the probability distribution of the training samples is not the same as the true underlying distribution \cite{fahse_managing_2021}. An example of this kind of bias is observed in the open-source image data set ImageNet. Shankar et al. \cite{shankar2017classification} shows how this data set doesn't have enough geographically diverse images that can have a broad representation across the changing world we live in.

    \paragraph{Label Bias}\label{lab-bias}
 \hfill\\
    Label bias or annotation bias arises when labels used to train the model are not entirely correct. This will lead to the training labels not representing the true labels. Training labels can systematically deviate as a result of vagueness and cultural or individual differences \cite{fahse_managing_2021}. 
    The authors in \cite{sap_risk_2019} investigate annotation bias in African American English (AAE) tweets. When annotators were primed to consider dialect, their assessment of an AAE tweet being "offensive" or "not offensive" was shown to be biased.

    \paragraph{Co-variate Shift}\label{co-bias}
 \hfill\\
    This kind of bias arises when the distribution of the features used to train the model is different in the training and testing phase. 
    Imagine a model used to invite job interviewees that is trained using skills demanded by the industry a few years ago. If this model was to be used now, it would not be able to make accurate predictions since the skills demanded by the industry have shifted significantly
    \cite{gu2019understanding}.
    
    \paragraph{Sampling Bias}\label{sampling-bias}
 \hfill\\
    Sampling bias can occur when sampling of subgroups is not random \cite{mehrabi_survey_2021}. An example of this kind of bias can arise if a researcher who's interested in learning the movie preferences of teens post on a social media group. This would primarily capture the preferences of teens who are members of that group. Sampling bias manifests here because teens who use social media and are part of the group might have different characteristics than teens who don't even have social media accounts. To conclude, the researcher might not get an accurate representation of the movie preferences of teens as a whole.
    
    \paragraph{Specification Bias}\label{spec-bias}
 \hfill\\
 This sort of bias arises during the specification of what comprises the input and output during a learning task \cite{ThomasH}. The specifications are often prepared by system designers. If the design choices are misaligned with end goals, they can exhibit specification bias \cite{tal_target_2023}.
 
 \paragraph{Aggregation Bias}\label{agg-bias}
 \hfill\\
 This bias arises when the true underlying patterns are misinterpreted by how the data is combined. Let's imagine there is a smartwatch for fitness that keeps track of the number of steps that you take constantly throughout the day. The numbers of steps are then aggregated into weekly averages and fed to train an ML model. Aggregating the data weekly can make the model miss vital information, like predicting if the person reached their daily step goals, which can only be obtained from the short-term patterns of the data.
 
    \paragraph{Linking Bias} \label{link-bias}
 \hfill\\
    Linking bias occurs when true behavior of users is misrepresented due to the attributes of networks derived from user interactions, connections or activity \cite{olteanu2019social}. Let's consider a social networking site where you connect with other users. The platform has some users who are highly active and have lots of connections, and others who are socially active in the physical world but not as active and have fewer connections on the platform. Linking bias will arise if the highly active users appear more central in the network due to their numerous connections. This could lead to an inaccurate representation of the actual social dynamic. 
    
    \paragraph{Inherited Bias}\label{inherited-bias} 
 \hfill\\
 According to \cite{ThomasH} inherited bias arises when biased output from a tool is used as input for other machine learning algorithms. In simple terms, the new algorithms can inherit the bias from tools' output. Let's assume the output of an automated loan approval tool is used as input to a new machine learning model. If the loan approval tool is biased regarding race, then this bias will be inherited by the new model.

    \paragraph{Longitudinal Data Fallacy}\label{log-bias}
 \hfill\\
This sort of bias can arise when cross-sectional analysis is performed to study temporal data. Let's consider a study done to analyze the sales performance of item A in all stores of a specific area over 2 months. The results showed that the sales performance of item A was poor in all stores. However, after examining the data over 36 months, it was found that the sales of the same item in the same stores improved over time. This shows that a fallacy can emerge when temporal dynamics are not considered; instead, cross-sectional analysis is used.

    \subsubsection{Human Bias}\label{hb}
 \hfill\\
 Human bias can be present in the machine learning pipeline in various stages. It can be present as prejudices in the training data. Bias can also be introduced by humans in the development and deployment stages.
 
    \paragraph{Historical Bias}\label{historical-bias}
    \hfill\\
Historical bias can occur if models perpetuate bias present in the historical record. These biases can arise from sources including social prejudices, preconceived notions and unequal treatment of different individuals or groups. The authors in \cite{ghosh_chatgpt_2023} found that ChatGPT associated the action of cooking breakfast to female entities when asked to translate a gender-neutral sentence from Bengali to English. This can be due to backdated cultural and societal expectations of women to take on domestic roles like cooking, cleaning, etc.

    \paragraph{Population Bias}\label{pop-bias}
 \hfill\\
This sort of bias can arise when there are biased variations in user characteristics or demographics between the target population and the population of users represented in a data set \cite{olteanu2019social}. Let's consider there is a mobile application to rent cars in a country. This application is primarily used by people living in the cities and not at all used by people residing in rural areas. If the data collected from this application is used to study the general preferences of the whole country's population, then it may exhibit bias towards people inhabiting the city.

    \paragraph{Self-selection Bias}\label{ss-bias}
 \hfill\\
    This kind of bias arises when subjects have the full right to decide if they want to participate in a study or not. Let's consider a shopping site, where some active users provide reviews of the products they purchase. These users are self-selecting to give reviews. There are some users who purchase items, but never post reviews. If a recommendation system is created using these self-selected user reviews, then it might miss the preferences of the users who didn't share their reviews. Thus, the system will be biased toward active users.
    
    \paragraph{Behavioral Bias}\label{bb}
 \hfill\\
Behavioral bias occurs when user behavior across platforms or contexts displays systematic distortions \cite{olteanu2019social}. Jiang et al. \cite{jiang_little_2016} discusses how users can find it difficult to transfer learning from one platform to another, especially if the platforms are not very similar. This implies the user needs to change their behavior according to the characteristics of each platform since they are not very similar.

    \paragraph{Temporal Shift}\label{ts}
 \hfill\\
    This kind of bias is described as the systematic distortions that arise across behaviors over time or among different user populations\cite{olteanu2019social}. The authors in \cite{gurjar_effect_2022} presented findings on how users start posting more content after they hit a popularity shock. Here, a temporal shift could include any changes in the users' engagement patterns that concur with their newfound popularity.
    
    \paragraph{Content Production Bias}\label{cpb}
 \hfill\\
    This bias occurs when user generated content has behavioral bias expressed as lexical, semantic, syntactic and structural differences \cite{olteanu2019social}. The authors in \cite{paris_differences_2012} show how different communities in social media exhibit significant differences in language use for the content posted.

    \paragraph{Deployment Bias}\label{dep-bias}
 \hfill\\
 Deployment bias emerges when a model is used or interpreted in a way that is not deemed appropriate when deployed to be used in the real world \cite{fahse_managing_2021}. This will occur when a machine learning model is built and evaluated, assuming to be fully autonomous when, in reality, it is used in a complex socio-technical environment that follows human decisions \cite{fahse_managing_2021}. As outlined by \cite{lee2021}, an insurance company that used a fraud detection model resolved to use a human investigator feedback loop. When a fraud prediction was made by the model, human investigators would review the cases to validate the prediction \cite{lee2021}. Deployment bias arises here because of how human validation interacts with the outcomes of the model. Humans checking the outcomes can unconsciously inject their biases into the validation process.
 
    \paragraph{Feedback Bias}\label{fee-bias}
 \hfill\\
 This kind of bias arises when the output of a model influences features or inputs that are used for retraining or refining the model \cite{fahse_managing_2021}. An example of this bias can occur if a movie recommendation system keeps suggesting movies based on the user's past choices, which in turn limits exposure to other diverse options. The bias arises because the model uses the user's choices to retrain itself to predict movies.
 
    \paragraph{Popularity Bias}\label{poty-bias}
 \hfill\\
 Popularity Bias arises when well-liked items get more exposure \cite{mehrabi_survey_2021}. The authors in \cite{sushma_channamsetty_recommender_nodate}, shows that recommender systems suggest items solely relying on popularity and not aligning with user preferences.
 
    \subsubsection{Model Bias} \label{mb}
    \hfill\\
    This section describes how model design choices can cause biased predictions.
    
    \paragraph{Algorithmic Bias} \label{alg-bias}
 \hfill\\
       This bias is caused by the algorithm itself and not the input data \cite{ricardo2018}. It could arise from structure, design and/or decision-making aspects of the algorithm itself. Let's consider there is a facial recognition model that was trained to be fair and unbiased. When evaluating this model, it was found that it consistently misidentified individuals from a certain ethnicity. This sort of bias was not present in the training data, so it is possible that the bias may stem from the internal mechanism of this model. The algorithm may process facial features from cultural backgrounds differently and unknowingly cause this issue.
       
       \paragraph{Evaluation Bias}\label{eva-bias}
 \hfill\\
 Evaluation bias emerges when the data used to test the model is not even close to what it would encounter in the real world. According to \cite{fahse_managing_2021}, if a wrong benchmark set is selected, then it can lead to neglecting potential bias. Let's consider a smile detector model trained using a data set without proper representations of Asian individuals. 
 When testing the model, if the benchmark used is unbalanced similar to the training set, then the bias against Asians will be overlooked.




\section{Practical cases of unfairness in real-world setting}
Unfairness can be present in ML models in many real-world scenarios. This can manifest across various platforms and domains. For the last few decades, the idea of using automated tools to aid decision-making processes in numerous societal contexts has been made popular. This increased use of these tools has raised the question of fairness in the predictions provided by them. Models like these are present in various sectors, the subsections below describe some of these sectors in more detail.

\subsection{Criminal Justice System}
The use of automated tools is now a norm in the criminal justice system. They are applied to various aspects of the justice system like law enforcement, corrections and court cases \cite{ail24,067d2e4389db4d9cba8865910a8cec71,manning_towards_2018}. The most commonly used automated models in criminal justice are for offense profiling and risk assessment\cite{ais24}. The most widely known case of biased prediction in criminal justice is COMPAS (Correctional Offender Management Profiling for Alternative Sanctions), which was used as a recidivism indicator. COMPAS was widely used to predict the likelihood of a person to re-offend within two years \cite{tim09}. The authors in \cite{mattu_machine_nodate} discuss how, despite having similar overall accuracy, the COMPAS algorithm discriminated against Black defendants by predicting a higher risk while favoring White defendants by assigning them a much lower risk. Similar to this case, legal researchers found that HART (Harm Assessment Risk Tool), which was used by the police to aid in decision making, prioritized assigning the high-risk individuals a low-risk score over assigning the low-risk individuals a high-risk score \cite{oswald18}. This, in turn, raised concerns about negative impacts on society from certain individuals who were given low-risk scores (but turned out to be high risks). Chouldechova \cite{chouldechova2016fair} describes how it is necessary to adjust error rates in predictors to achieve fairness when predicting recidivism scores across different groups. Racial bias in forensic databases can also impact automated tools' contributions to the criminal justice system \cite{risher}. Gstrein et al. \cite{067d2e4389db4d9cba8865910a8cec71} states how using black-box models to predict recidivism scores, guilt or innocence of convicts in court, lacks conformity with legal regulations since it raises concerns about transparency, fairness and following legal principles in general.

\subsection{Hiring Employees}
Machine learning models have been utilized by a growing number of organizations to aid in making employment decisions, but it's not without any challenges. If not evaluated properly, a seemingly fair model will perpetuate bias against certain groups when hiring predictions are made, especially if it is trained on historical data \cite{barocas_big_2016,ais24}. The authors in \cite{Hanna_2020} found gender and racial biases in Facebook's advertisement delivery system for employment and housing advertisements. They also discuss how the delivery system used for Facebook advertisements can alter the actual audience based on the advertisement's content. The authors of the papers \cite{sweeny13,datta2015automated} discuss their findings on how Google's advertising system was biased against certain subgroups when employment advertisements were made on the platform. Various research shows that major employers discriminate against females and some ethnic groups \cite{Bendick,Johnson}. The authors in \cite{rag} tested 18 vendors which used automated hiring models and found only a mere seven addressed bias in their hiring model.

\subsection{Finance}
The financial industry increasingly employs automated tools to aid decision-making processes. However, their use of such tools comes with challenges that include ensuring compliance with consumer laws and minimizing disparate impacts for disadvantaged communities and making fair predictions in general. Sigalos et al. \cite{sigalos_i_2023} presents several cases of bias exacerbated by models when lending loans for fraud detection and investment recommendation.
In 2009, the author in \cite{lieber_american_2009} initiated a debate about fair algorithmic models used in fin-tech. They mentioned how American Express potentially used historical data to predict the future behaviors of current users. Various companies that use historical data to train their models are vulnerable to getting a biased model. There has been a lot of research that shows historical bias against Hispanic, Black and some minority communities for creditworthiness and interest rates \cite{butler2020racial,fuster2022predictably}. Some credit card companies use "financial profiling" to judge a candidate's creditworthiness \cite{news_gma_nodate}. In this case, the bias arose when the prediction was based not only on the candidate's financial behavior but also their shopping habits from certain stores. As transparency in these models has become a legal obligation for financial institutions, they are vulnerable to disparate impact \cite{fino21,BARTLETT202230}.

\subsection{Healthcare}
Discrimination in AI models used in healthcare is an increasing concern since it has serious consequences. Bias in such models can lead to inequitable practices and inaccurate diagnoses. Straw et al. \cite{straw22} demonstrated the presence of gender bias in a model that is used for diagnosing liver disease. A similar case emerged when an X-ray prediction model was shown to exhibit bias, when it wrongly predicted patients from certain groups like women, Black, Hispanic and young individuals to not need medical attention \cite{cho_rising_2021}. Numerous other cases of biased models were shown to discriminate against certain ethnic groups \cite{ruha19,yogarajan_data_2022,noauthor_individualized_nodate,obermeyer_dissecting_2019}. Presumably, biased algorithms have been developed for tasks like heart surgery, kidney transplants, rectal and breast cancer, which seldom affected access to services and resource allocation \cite{obermeyer_dissecting_2019}. The lack of inclusive datasets used to train models utilized in healthcare can add to the systematic under-representation \cite{celi_sources_2022,benjamens2020state}.

\subsection{Education}
Educational institutions employ machine learning models to achieve various goals, including setting the curriculum for students, deciding what resources they should receive and other critical decisions \cite{ais24}. These models have the potential to manifest bias in various ways. The scores given by e-rater (an automated essay scoring system) had some discrepancies with human-assigned scores, especially for some Asian individuals \cite{bridgeman2009considering,bridgeman2012}.  
 During COVID, students were awarded grades using ML models since they couldn't appear for tests physically. The models used to score the students were shown to discriminate based on certain sensitive attributes \cite{adams_england_2020,lee_factcheck_2020,denes_case_2023}. The authors in \cite{baker2021algorithmic} presented a thorough survey about biased models used in education; those interested can check its content.

\subsection{Others}
Bias can creep into several systems that are used in our daily lives. A popular ride-hailing service, Uber, was found to have a customer rating system that was a channel to workplace discrimination against drivers from certain ethnic backgrounds \cite{rosenblat_discriminating_2017}. ChatGPT displayed bias against women when asked to translate sentences in certain languages \cite{ghosh_chatgpt_2023}. Bias was also detected in facial recognition systems like Amazon  Rekognition, IBM Watson,  Microsoft Azure face API and Google cloud vision API \cite{wen_phenomenological_2023}. Bias in advertisement delivery is also a potential issue that can affect how users are impacted by the use of AI models. According to the authors in \cite{noauthor_study_2021}, the Facebook advertising algorithm was shown to display gender bias since men were more likely to get ads for jobs that are in male-dominated fields, and women were more likely to get ads that are in female-dominated domains. Bias in AI systems used in gaming is also reported as the form of discrimination against atypical behavior (for example, an autistic player) \cite{noauthor_autistic_2011} and training on data from hardcore-player (who doesn't represent the whole gaming population) \cite{melhart2023ethics}. Airbnb's facial recognition system wasn't able to match an Australian's (with South-Asian heritage) selfie with the government-issued ID and sparked potential racial bias \cite{iqbal_airbnbs_2023}. This highlights the issue of bias against certain populations who might be under-represented in the training data. In January this year, the Nine network used an image of a female Australian politician in a news broadcast, and it was discovered that the image was digitally altered \cite{noauthor_it_2024}. They stated it was an unintended consequence of the AI resizer used. In 2015, Google's photo service was blamed for labeling a photo of a Black individual as a gorilla, and the solution Google came up with after two years was to remove the word "gorilla" from the labels used for pictures \cite{simonite_when_nodate}. The cases mentioned here are noteworthy, but covering all the different sectors that fell prey to biased results from AI models is out of the scope of the paper.

\section{Ways to mitigate bias and promote Fairness}
It is crucial to remove bias from ML models, especially to make them responsible and ethical. Models can contain historical bias and discriminate against certain groups or individuals. To align with the principles of social justice, we need the models to be fair. Addressing bias can make the system more inclusive since it considers a diverse environment. Also, a model behaving unfairly will make the user less likely to accept them. Users will be able to trust and use fair and transparent systems with more confidence. Additionally, responsible systems will ensure that all laws and regulations related to fair and non-discriminatory practices are adhered to.

Mitigating bias in ML models can be a complex task that requires an amalgamation of technical, ethical, explainable and organizational strategies. It is vital that these issues are addressed at different stages of the ML pipeline.

\subsection{General Stages to mitigate bias}
\begin{figure}[h]
    \begin{overpic}[
      width=0.8\linewidth]{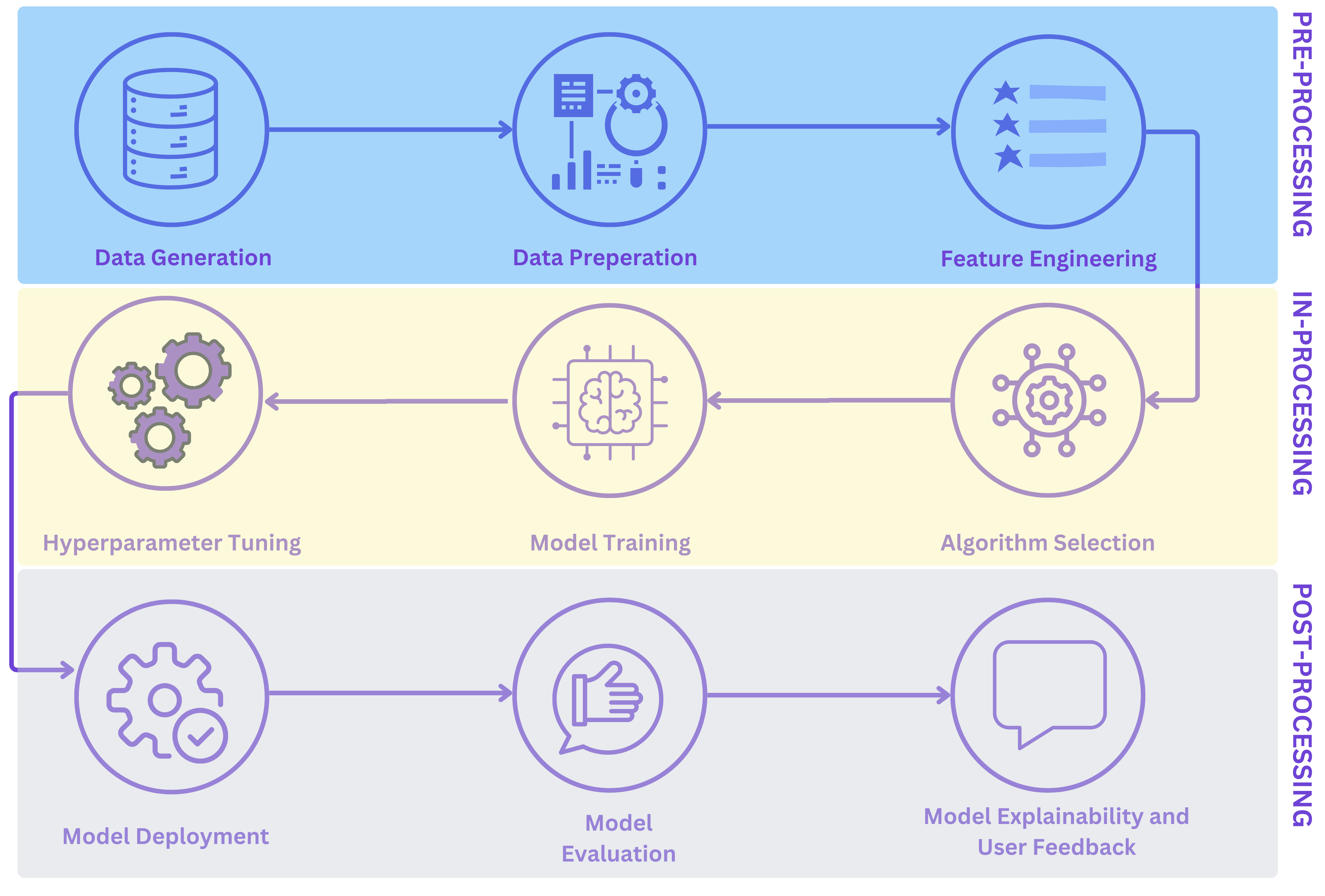}
    \end{overpic}
  \caption{The three crucial stages in the development of an ML model}
        \label{fig:ml-bias-stage}
\end{figure}

Recently, extensive research has been conducted to promote fairness in machine learning models. These bias mitigation strategies can generally be categorized into three types: pre-processing, in-processing and post-processing. Figure \ref{fig:ml-bias-stage} shows these three stages in the ML pipeline.

\subsubsection{Pre-processing}
\hfill\\
Pre-processing strategies help remove bias from training data. These approaches try to ensure unfair patterns are reduced or removed from training data before it is used to feed the model. The impartiality and quality of training data influence the model's effectiveness in making fair predictions \cite{farayola23}.

\subsubsection{In-processing}
\hfill\\
This approach mitigates unfairness in the ML models training process and aims for bias-free outcomes. This strategy focuses on properly tuning and developing the algorithms used \cite{farayola23}. According to the authors in \cite{farayola23}, choosing an algorithm that is less vulnerable to bias and a proper hyper-parameter tuning process is vital during this stage.

\subsubsection{Post-processing}
\hfill\\
If a model's actual outcome is unfair with respect to one or more sensitive attributes, this strategy can mitigate that by modifying the model's predictions. This strategy can be employed only when the model is done training and has made its initial predictions.
\hfill\\

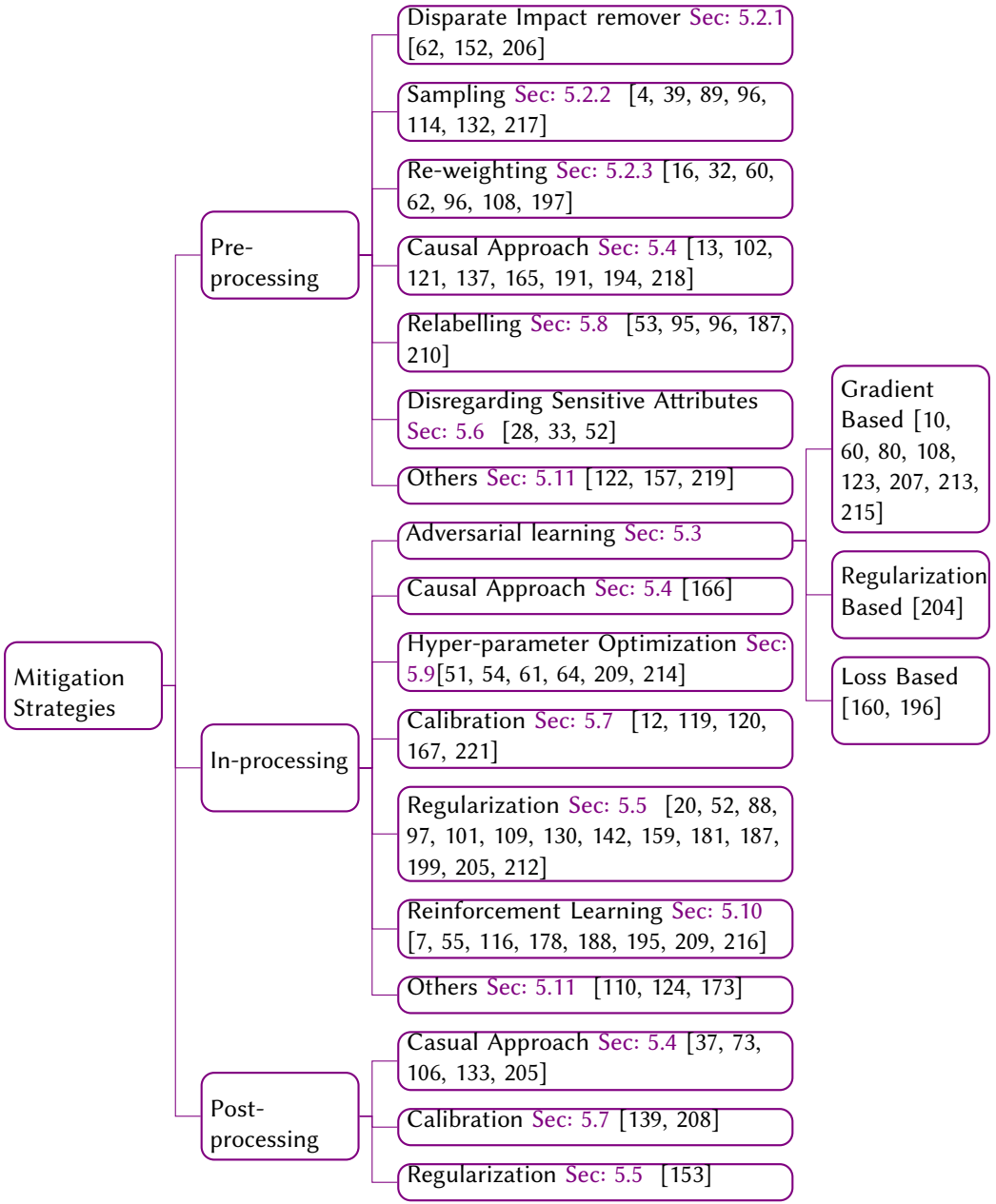
\begin{figure}[ht]
    \begin{forest}
    for tree={
        my node,
        parent anchor= east,
        grow' = east,
        draw=violet, 
        child anchor = west,
        l sep = 15pt,
        rounded corners=5pt,
        edge path={
            \noexpand\path[\forestoption{edge}]
            (!u.parent anchor) -- +(5pt,0) |- (.child anchor)\forestoption{edge label} [violet];
        },
    },
    [Mitigation Strategies,text depth=0ex, minimum width=0.2cm
        [Pre-processing , minimum width=0.5cm
            [Disparate Impact remover {\textcolor{violet}{Sec: \ref{dir} }}\cite{feldman2015certifying,pess24,xu2020removing} , text depth=1.9ex, minimum height=0.8cm,text width=35ex]
            [Sampling {\textcolor{violet}{Sec: \ref{samp} }} \cite{miron21,Sabina21,agarwal2019fair,chouldechova2016fair,kamiran_data_2012,zhang_framework_2023,iosifidis2020fae} , text depth=1.9ex, minimum height=0.8cm,text width=35ex]
            [Re-weighting {\textcolor{violet}{Sec: \ref{re}}}  \cite{feldman2015certifying,Chakraborty_2020,fav23,Bellamy19,lahoti2020fairness,kamiran_data_2012,wang_analyzing_2021}, , text depth=1.9ex, minimum height=0.8cm,text width=35ex]
            [Causal Approach {\textcolor{violet}{Sec: \ref{ca} }}\cite{loftus2018causal,vig2020causal,Bareinboim,NIPS2017_1271a702,Nabi_Shpitser_2018,vanbreugel2021decaf,kilbertus2018avoiding,zhang2023causal}, text depth=1.9ex, minimum height= 0.8cm,text width=35ex]
            [Relabelling {\textcolor{violet}{Sec: \ref{rel} }} \cite{Dunkelau2020FairnessAwareML,kamiran_data_2012,kamiran09,tavakol2020fair,yang23}, text depth=2ex, minimum height= 0.8cm,text width=35ex]
            [Disregarding Sensitive Attributes {\textcolor{violet}{Sec: \ref{dsa} }} \cite{chaudhary2023practical,Calders10,du2021fairness},, text depth=2ex, minimum height= 0.8cm,text width=35ex]
            [Others {\textcolor{violet}{Sec: \ref{othe}}} \cite{luong11,Zhang_Hernandez-Boussard_Weiss_2023,Qraitem_2023_CVPR}, text depth=0ex, minimum height= 0.5cm,text width=35ex]
        ] 
        [In-processing 
            [Adversarial learning {\textcolor{violet}{Sec: \ref{adv}}} , ,text depth=0ex, minimum height= 0.5cm, text width=35ex
            [Gradient Based \cite{lahoti2020fairness,zhang2018mitigating,ball21,grari_fairness-aware_2020,Zafar_2017,fav23,xu2021robust,Ma_2023}, text depth=9.5ex, minimum height = 2.3cm]
            [Regularization Based \cite{wu2021fairrec}, text depth=2ex]
            [Loss Based \cite{wang_balanced_2019,reimers2021learning}, text depth=3ex]
            ]
            [Causal Approach {\textcolor{violet}{Sec: \ref{ca} }}\cite{salimi19}, minimum height= 0.5cm,text depth=0ex, text width=35ex]
            [Hyper-parameter Optimization {\textcolor{violet}{Sec: \ref{hyp}}}\cite{inbook,dooley2023rethinking,jenny23,F_Cruz_2021,fetterman2023tune,zafar2017fairness}, text depth=2ex, minimum height= 0.8cm,text width=35ex]
            [Calibration {\textcolor{violet}{Sec: \ref{cali} }} \cite{liu2019implicit,liu2017calibrated,hébertjohnson2018calibration,ocaa283,Salvador2021BiasMO}, minimum height=0.8cm,text depth=2ex, text width=35ex]
            [Regularization {\textcolor{violet}{Sec: \ref{reg} }} \cite{Olfat20,wang2021enhancing,Lahoti_2019,tavakol2020fair,Kamishima12,BerkHJJKMNR17,sundararaman2022debiasing,memarrast2021fairness,yu2022policy,wu2018fairnessaware,raman2021data,hu21,khedr2022certifair,du2021fairness}, text depth=3.8ex, minimum height=1.3cm, text width=35ex]
            [Reinforcement Learning {\textcolor{violet}{Sec: \ref{rein} }}\cite{Timmaraju_2023,atwood2019fair,li_hiring_2020,jenny23,wang2019mitigate, elmalaki2021fair,zhang2021recommendation,Singh2021BuildingHR},text depth=2ex, minimum height= 0.8cm,text width=35ex]
            [Others {\textcolor{violet}{Sec: \ref{othe} }} \cite{omid23,lassig2022metrics,mahabadi2020endtoend}, text depth=0ex, minimum height= 0.5cm,text width=35ex]
        ]
        [Post-processing 
            [Casual Approach {\textcolor{violet}{Sec: \ref{ca}}}  \cite{kusner_counterfactual_2018,Galhotra_2017,misher2021,wu2018fairnessaware,chiappa2018pathspecific}, text depth=2ex, text width=35ex, minimum height =0.8cm]
            [Calibration {\textcolor{violet}{Sec: \ref{cali}}} \cite{ziqi23,campero19} , minimum height= 0.5cm,text depth=0ex, text width=35ex]
            [Regularization {\textcolor{violet}{Sec: \ref{reg} }} \cite{petersen2021postprocessing},text depth=0ex, minimum height= 0.5cm, text width=35ex]
        ] 
    ]
    \end{forest}
     \caption{Proposed Taxonomy of the Mitigating Strategies}
        \label{fig:ml-ms}
\end{figure}

It is important to realize that each of these three strategies has its own limitations. A single strategy alone may not be enough to tackle bias. So, it is important to decide which mechanism is best by taking under consideration the nature of the dataset, type of bias, fairness metric used and chosen model characteristics. It is also important to realize that some of these techniques can end up reducing accuracy \cite{kleinberg2016inherent}, so trade-offs between accuracy and fairness should be evaluated. Figure \ref{fig:ml-ms}, shows a taxonomy of all the mitigation strategies identified in the current literature.

\tikzset{
    my node/.style={
        thick,
        minimum height=1.2cm,
        minimum width=1.2cm,
        text width=13ex,
        text height=0ex,
        text depth=0ex,
        font=\sffamily,
    }
}

\subsection{Removing Bias from Data}
De-biasing the data used to train a model is the first crucial step towards designing a fair and ethical model. It not only promotes fairness but also helps increase the accuracy of the model since it considers a more diverse dataset and, in turn, produces a more generalized outcome. Having a dataset without bias will lead to a responsible and inclusive model that will be useful for all members of our society.

\subsubsection{Disparate Impact remover} \label{dir}
\hfill\\
This process involves manipulating and transforming data to address fairness. Initially, the potential disparate impacts on the sensitive attributes of the dataset are examined \cite{feldman2015certifying}. Next, the dataset is \textit{repaired} by transforming the initial dataset to fix the issues identified. During this stage, rank is preserved to ensure the relative position of data points is not altered.  The main aim of the approach is to address disparate impacts, which is important in reducing unintended consequences.The authors in \cite{feldman2015certifying} does this by modifying the attributes in the dataset to ensure the distribution of protected and unprotected groups are closer. The authors in \cite{xu2020removing} take a different path, and use a modified version of  differentially private stochastic gradient descent (DPSGD) to remove disparate impact. 

\subsubsection{Sampling} \label{samp}
\hfill\\Sampling techniques can help address data bias by considering data misclassification and class imbalance. According to \cite{miron21}, data sampling can help effectively analyze data without losing its universality. Feature sampling is the process of selecting only relevant features for the model to train on and can also contribute to mitigating bias \cite{miron21}. Kamiran et al. \cite{kamiran_data_2012} describes two approaches of sampling, which include:
\begin{itemize}
    \item Uniform Sampling: promotes fair dataset by replacing objects according to their respective weights.
    \item Preferential Sampling: mitigates discrimination by resampling data too close to the decision boundary.
\end{itemize}
Many papers, including \cite{kamiran_data_2012,Sabina21,agarwal2019fair,chouldechova2016fair}, incorporate some form of sampling to ensure that the dataset used does not discriminate against any sensitive attribute groups. An interesting approach is taken by the authors in \cite{wang23}, where they randomly remove training examples that are related to the over-represented demographic groups and add samples to the under-represented groups using re-sampling.
\subsubsection{Re-weighting}\label{re}
\hfill\\
Similar to resampling, re-weighting is a pre-processing technique that can be used to promote fairness in datasets. This process simply assigns different weights to different data objects based on their relevance. This method is widely used for models that have the opportunity of weighted samples like SVMs. Kamiran et al. \cite{kamiran_data_2012} describes the weighting process, which efficiently up-weights data that are disadvantaged and down-weights data that can cause discrimination. According to \cite{wang_analyzing_2021}, bias caused by missing values can be addressed using re-weighting techniques. This method can improve fairness, although there is a slight decrease in accuracy \cite{wang_analyzing_2021}. Several studies describe how re-weighting techniques can be applied to address issues like selection bias and disparate impact discrimination \cite{feldman2015certifying,Chakraborty_2020,fav23,Bellamy19}. The authors in \cite{lahoti2020fairness}, take a different approach by integrating adversarial learning and re-weighting to improve the fairness of a model with unobserved sensitive attributes in the training data. The adversarial reweighed learning process mentioned by \cite{lahoti2020fairness} includes the components below:
\begin{itemize}
    \item The main classification model
    \item An adversary for finding regions with high error rates
    \item Weights which are assigned by examining the error rates. A higher weight is assigned to error-prone regions to encourage the model to focus on getting better at being fair.
\end{itemize}
This process helps encourage the model to improve predictions for all sensitive groups without knowing them explicitly.

\subsection{Adversarial Learning}\label{adv}
Adversarial learning can be employed to mitigate bias in ML models indirectly. There are several ways in which adversarial learning can be used to de-bias. In \cite{zhang2018mitigating}, the authors describe how the process of \textbf{gradient-based adversarial debasing} can help mitigate bias in models. The process in \cite{zhang2018mitigating} can be summarized as:
\begin{itemize}
    \item A predictor is used to get outcomes given $X$ (which is the main task)
    \item An adversary tries to predict the sensitive variable $S$, based on the prediction made by the predictor.
\end{itemize}
The predictor is trained in a way that not only fulfills its primary task but also maintains fairness by actively counteracting the effect of the adversary\cite{zhang2018mitigating}. The authors in  \cite{ball21,grari_fairness-aware_2020,fav23} employ this type of adversarial de-biasing to promote fairness in their models.
\hfill\\
A \textbf{loss-based adversarial de-biasing} method will involve the process of optimizing the loss functions as well as the main task (main classification task) and auxiliary task (adversary for predicting sensitive attribute). The authors in \cite{wang_balanced_2019} implements two competing loss functions:
\begin{itemize}
    \item Classifier loss that motivates the classifier to perform well.
    \item Critic loss which penalizes if it's not able to predict the sensitive attribute from the classifier's outcome.
\end{itemize}
In \cite{wu2021fairrec}, authors employ a \textbf{regularization-based adversarial de-biasing} method. Their model includes:
\begin{itemize}
    \item A predictor (which is the main model).
    \item An attribute discriminator that predicts sensitive attributes based on model predictions.
    \item An orthogonality regularization term that motivates bias-free embedding without correlations.
\end{itemize}

Yang et al. \cite{yang23}, describe how adversarial learning can be used to disentangle non-sensitive and sensitive features by optimizing Balanced Fairness Objective (BFO) for a recommendation model. Their process combines both loss functions and min-max optimization to achieve fairness in the model.

\subsection{Causal Approaches}\label{ca}
For two random variables $A$ and $B$, causality is the phenomenon when $A$ causes $B$ \cite{loftus2018causal}. Making changes to $A$ will lead to a different outcome for the variable $B$. In simple terms, causality describes the relationship between two events, where one event directly causes the other event. Understanding causal relationships between features is essential for tackling bias since sensitive factors that are out of an individual's control should not be used to make decisions \cite{loftus2018causal}. The process of using causal relationships to mitigate bias can be summarized as \cite{zhang2023causal,vig2020causal}:
\begin{itemize}
    \item Identify and understand the causal pathways through which sensitive attributes can influence the model's outcome. This will help figure out which levers can be adjusted.
    \item To disrupt the causal pathways, design interventions to reduce the bias caused by them. For this, techniques such as neuron adjustments, data augmentation, etc., can be used.
    \item To avoid any unintentional results from modifying the model, use only fine-grained adjustments, not anything drastic.
\end{itemize}
Several studies have been conducted to obtain a fair model by using causal reasoning, including \cite{Nabi_Shpitser_2018,NIPS2017_1271a702,salimi19,kilbertus2018avoiding}. In \cite{Bareinboim}, the authors merge three different causal analysis methods to resolve selection bias and confounding. Galhotra et al. \shortcite{Galhotra_2017} introduces a novel causality-based metric to measure dissemination and uses it for software fairness testing. Counterfactual reasoning is an approach used to answer what would have happened if some factor was different (e.g. if the outcome would be the same for a person from a different race). Causal learning can be utilized to identify counterfactuals. The authors in \cite{kusner_counterfactual_2018} describe how a fair predictive model can be created by restricting the outcome to be impacted by any sensitive variable (which is identified using a causal graph) and using counterfactual reasoning to ensure hypothetical interventions are considered.

\subsection{Regularization}\label{reg}
Regularization helps make the ML model more general. It adds constraints and penalties in the loss function to discourage unfair outcomes. The main objective of this technique is to prevent overfitting and create simpler models. The authors in \cite{wang2021enhancing} constructs and optimizes a custom loss function by using a novel regularization term to introduce causality during the training process. Regularization has also been utilized to satisfy multiple definitions of fairness \cite{kang2021multifair,tavakol2020fair}. Kamishima et al.  \cite{Kamishima12} introduced a prejudice remover, which penalizes the model if there are high correlations between the target and sensitive variables. Utilizing regularization for promoting fairness has significant potential and has been used in a de-biasing technique \cite{sundararaman2022debiasing}, fairness-aware ranking system \cite{memarrast2021fairness}, mitigating bias in reinforcement-learning agents \cite{yu2022policy} and enforcing fairness in deep learning models \cite{Olfat20}. Regularization is traditionally considered an in-processing technique, but can also be used in post-processing settings. Peterson et al. \cite{petersen2021postprocessing} utilized regularization to adjust the outputs of an existing model to ensure predictions for similar individuals are consistent.

\subsection{Disregarding sensitive attribute}\label{dsa}
It is important to ensure that the model doesn't learn to associate sensitive attributes like gender, race, age etc. when producing results. Some existing literature has taken the approach of removing the sensitive variables completely. This has been achieved by methods like adversarial learning \cite{poulain_improving_2023}. Various research has found ways to reduce or eliminate correlation between sensitive attributes and other attributes to mitigate any discrimination caused by them \cite{Calders10, gitiaux2021fair,du2021fairness,woodworth2017learning,creager2019flexibly}.
Although masking sensitive attributes seems like a viable solution to stop the model from forming association with them, there might be correlated features in the data that indirectly contain information about the sensitive attributes \cite{Chakraborty_2020}. So the technique of masking can prove to be ineffective against bias mitigation \cite{klein18}.

\subsection{Calibration}\label{cali}
Calibration can be described as the process of ensuring the proportions of positive predictions and proportions of positive examples are equal \cite{dawid82}. These techniques try to adjust the predictions of models to better reflect the true outcomes. There are two types of calibration techniques \cite{ziqi23}: 
\begin{itemize}
    \item Platt Scaling : Model's raw scores are adjusted to represent probabilities more reliably.
    \item Isotonic Regression : Regardless of any bias in the model, if it predicts something to be likely then it should actually happen more often.
\end{itemize}
The overall equality of positive predictions is desirable, but a well-calibrated model should maintain equal proportions of probability within different subgroups (eg. age, gender etc.). Fairness promoting calibration has been integrated by various researchers for tasks that include  accurately reflecting scores for loan applications and risk assessments  
 \cite{cynthia16,chouldechova2016fair,liu2019implicit,campero19} and preventing discriminated decisions in MAB (multi-armed bandit) setting \cite{liu2017calibrated}. The authors in \cite{hébertjohnson2018calibration} discuss an approach called multi-calibration, which goes beyond just calibrating probabilities for specific groups but targets certain subpopulations to mitigate bias and promote fairness and generalization. 

As mentioned in \cite{kleinberg2016inherent,pleiss2017fairness}, except for some exceptions (constrained cases) it is nearly impossible for a calibrated model also to satisfy equalized odds. Authors in \cite{pleiss2017fairness} suggest prioritizing either calibration or error rate (for fairness) instead of trying to achieve a balance of both.

\subsection{Relabelling}\label{rel}
This technique aims to modify the ground truth values in the training dataset, to ensure fairness notions are satisfied \cite{Dunkelau2020FairnessAwareML}. Data massaging \cite{kamiran_data_2012,kamiran09}, is the process of taking a number of training data and changing their respective ground truths. According to \cite{Dunkelau2020FairnessAwareML}, data massaging allows any classifier to learn on a dataset that is fair and promotes group fairness. The process mentioned in \cite{kamiran09,kamiran_data_2012} for modifying labels to remove bias can be summarized as:
\begin{itemize}
    \item A ranker to approximate the probability of the data points in the target class, without the sensitive attributes.
    \item Identifying the promotion candidates (candidates that are in -ve class but will be moved to +ve class) and demotion candidates (candidates that are in +ve class but will be moved to -ve class)
    \item Modify labels of an equal number of candidates by swapping them.
    \item Iteratively perform this modification process until a desired level of bias has been reduced.
\end{itemize}

Numerous research uses re-labelling to promote fairness, including \cite{tavakol2020fair,kamiran_data_2012,yang23}.
It is important to note that re-labelling does come with extra cost (finding distance for the data) \cite{Chakraborty_2020}, so this is recommended when removing data points entirely will affect the model.

\subsection{Hyper-parameter optimization}\label{hyp}
This strategy can be used not only to increase a model's performance but also to mitigate bias. This can be done by considering fairness criteria during the hyper-parameter tuning process. Dooley et al. \cite{dooley2023rethinking} has conducted research to prove that architectures and hyper-parameters can have a notable impact on fairness in models. The authors in \cite{jenny23}, use hyper-parameter-optimization (specifically grid search and five-fold cross validation), to find hyper-parameters values that not only increase the accuracy of the model, but also considers the fairness of the model. Finding the correct balance of accuracy and fairness is key when mitigating bias using hyper-parameter optimization, also known as fairness-aware hyper-parameter optimization (FHO). Cruz et al. \cite{F_Cruz_2021} defines accuracy and fairness as a MOO (Multi Objective Optimization) problem. This objective can be satisfied by using the hyper-parameters tuning method to find a collection of hyper-parameters that optimize fairness by using a weighted-scalarization process \cite{F_Cruz_2021} and Pareto methods \cite{fetterman2023tune}.

\subsection{Fairness through Reinforcement Learning}\label{rein}
Reinforcement Learning (RL) provides a unique way to mitigate bias in ML systems. This is a promising approach compared to supervised learning, especially in dynamic settings, due to its ability to learn from interacting with the environment. RL models can employ continuous learning because the model itself can adapt in real-time which can help mitigate bias by considering concept drifts and frequent data changes. Li et al. \cite{li_hiring_2020} utilized RL to explore and learn about demographic groups that are under-represented for a model to predict if a candidate is worth hiring or not. RL has also been used in improving public health strategies, including mitigating bias in precision contagion policies \cite{atwood2019fair} and mitigating bias perpetuated from data for predicting if a patient is infected with COVID-19 \cite{jenny23}. The authors in \cite{wang2019mitigate} utilize RL to promote fairness of a facial recognition system by using the margin that is used in the loss functions for different racial groups. Both \cite{zhang2021recommendation} and \cite{Singh2021BuildingHR} describe their shared focus on mitigating long-term biases from recommendation algorithms by using Reinforcement learning. RLHF (Reinforcement Learning from Human Feedback), where models adapt their behavior according to the feedback given by humans, has recently gained popularity in the quest of fair AI. The authors in \cite{ouyang2022training} and \cite{elmalaki2021fair} use human feedback as a guide to developing the models, making them more efficient and adaptive.
 The survey \cite{gajane2022survey}, offers valuable insight on fair reinforcement learning.

 \subsection{Others}\label{othe}
This section describes processes used to mitigate bias that don't fall under the umbrella of the previously mentioned sections but are worth mentioning. The authors in \cite{omid23}, describes an elaborate method to mitigate intersectional bias by employing techniques which include quantum computing alongside data augmentation, fairness-aware fine-tuning and RL. K-NN has also been used to mitigate bias for doing tasks like discovering potential discrimination in datasets \cite{luong11} and to identify local regions for a locally fair ensemble model \cite{lassig2022metrics}. The authors in \cite{Zhang_Hernandez-Boussard_Weiss_2023} propose a novel approach to promote fairness in survival prediction models, by using censorship and ranking consistency. Qraitem et al. \cite{Qraitem_2023_CVPR}, introduce a new method called "bias mimicking" that trains a model using multiple sub-samples. Within each class in the dataset this process mimics bias distribution of other classes, which helps reduce correlations with sensitive attributes. Mahabadi et al. \cite{mahabadi2020endtoend} proposes a novel approach to mitigating bias in NLP by having a secondary bias-aware model with the main NLP model. The bias-aware model's predictions are used to adjust the loss function of the main model. Similar to this technique \cite{jin2020efficiently} also uses multiple models to promote fairness. They introduce the UBM (Upstream Bias Mitigation) framework that uses transfer learning to address bias in models. An upstream model is bias-mitigated and then used to pre-train a downstream model.



\section{How Users can be affected by unfair ML Systems}
Whenever AI models are created, the developers mostly think about how they can be used to achieve a certain purpose. 
Usually the developers forget to think about what it will be like when using an AI model. They forget to cater for a good user experience. Bias in AI systems can profoundly impact users in several ways. It has various negative impacts like discrimination, limited opportunities, decline in reliance and privacy concerns. One of the most commonly used AI systems, ChatGPT, is susceptible to be a weapon of mass deception and can spread misinformation and create deep fakes \cite{sison2023chatgpt}. The use of ChatGPT or similar models as WMD (weapon of mass deception) can have many direct and indirect negative impacts on users, including erosion of trust and psychological manipulations. The authors in \cite{lew_ai_2020} state that truly good user experiences (UX) are engaging, fun and addictive. The goal should be to develop systems that evoke good emotions and go beyond just user satisfaction \cite{lew_ai_2020}.

\subsection{Prejudiced Model against UX}
Models are susceptible to negativity bias because users tend to put more weight on any negative interactions with a model over neutral or positive experiences \cite{experience_negativity_nodate}. The authors in \cite{chen2023} conducted research on how bias in recommender systems can impact user experience. They point out that although these kinds of systems perform well when controlled tests are performed on them, they can lead to frustrated users by suggesting irrelevant and/or unfair recommendations. Users can stop trusting AI systems entirely if they consistently provide predictions that are biased and inappropriate\cite{chen2023,zvi2004}. Moreover, certain model's biased predictions can lead to the user not being able to explore diverse options and limit their ability to find new items \cite{chen2023}. The authors in \cite{siberstein20} discuss how models that are used for ad selection could lead to biased or irrelevant ads, which in turn triggers users to close them due to their negativity, ultimately leading to a not-so-pleasant experience. Popularity bias, mentioned in section \ref{poty-bias}, can lead to user dissatisfaction since they only get to interact with "popular" posts \cite{lacic2022drives}.
Targeting good UX requires mitigating bias in AI models. By ameliorating bias from models, we can hope to develop a user experience that is not only enjoyable but also fair.
\subsection{Ethical Considerations}
The idea of AI systems either matching or surpassing human capabilities is a possibility that can only be controlled through the implementation of solid moral standards \cite{gordon21}. So, when developing and deploying such systems the designers need to carefully consider the numerous ethical concerns. Currently, there are no universally enforced laws outlining ethical considerations for AI models. But there are significant development in designing such guidelines and policies to serve as recommendations or best practices by developers and organizations. UNESCO has recommended a set of ethical guidelines, which follow core principles like fairness, transparency and sustainability \cite{noauthor_ethics_nodate}. The European Union has their own regional regulations for AI applications and proposes ways to develop fair and transparent models \cite{noauthor_ethics_nodaate}. Several countries are working to create their own ethical guidelines for AI models, including Australia \cite{resources_australias_2022}, Singapore \cite{noauthor_pdpc_nodate} and Canada \cite{secretariat_responsible_2018}.
The authors in \cite{jobin_global_2019}, explore the existing ethical guidelines that are available for AI worldwide. They found some key principles that are considered globally. These key principles include

    \subsubsection{Transparency to promote Explainability}
    \hfill\\
    The goal of this concept is to make AI systems more understandable and interpretable \cite{adadi18}. The authors in \cite{ALI2023101805} break down the concept of XAI (Explainable Artificial Intelligence) into 4 sub-concepts.
    \begin{itemize}
        \item  Data Explainability: Understanding if there are biases that are perpetuating from the data used to train the model is essential. The designers should make the data collection and preparation process transparent so that the end users can understand it.
        \item  Model Explainability: This is to ensure the internal details of the model are transparent and understandable to users.
        \item Post-hoc Explainability: This is used to explain an outcome that is already given by a complex model. This will help get an idea of the model's reasoning but might not be able to reflect on the big picture.
        \item Explanation Assessment: This process aims to evaluate how well the XAI methods work in explaining predictions. 
    \end{itemize}
  This concept is used for various reasons including to minimize harm and improve AI modes
    \cite{an}, to comply with legal regulations \cite{flo18} and to gain user trust \cite{ag}.
   
    \subsubsection {Justice, Fairness and Equity}
    \hfill\\
    Justice, fairness and equity are very complex concepts, especially in the realm of AI systems. The three concepts can be summarized as follows \cite{jobin_global_2019}:
    \begin{itemize}
        \item Justice: This concept is linked with following rules and putting a stop to bias and discrimination. It also includes ideas about inclusion, diversity and the right to appeal any unfair decisions.
        \item Fairness: This concept ensures outcomes of AI systems are unbiased and addresses its impact on society.
        \item Equity: This phenomenon ensures that an AI system is designed to predict outcomes that are equitable for every individual.
    \end{itemize}

    \subsubsection {Non-maleficence}
    \hfill\\
    Non-maleficence upholds the principle of "do not harm"\cite{Floridi2019Unified}. Another intertwined concept is that of beneficence which is related to the principle of "do only good". Although these two principles sound quite similar, beneficence triggers an action, whereas non-maleficence may prompt not to \cite{werthner_introduction_2024}. For instance, if this concept was to be applied to health care,
    beneficence would require a model to predict diagnosis for helping the patient and non-maleficence would require the model to avoid causing harm to the patient (like prescribing the wrong dosage of a medicine).
    
    \subsubsection {Responsibility and Accountability}
    \hfill\\
    Responsible AI ensures AI is developed, assessed and deployed in a safe and ethical way \cite{mesameki_what_2024}. The authors in \cite{dignum2019responsible} extends this definition and states responsible AI doesn't only ensure systems are developed in a good way but also for a good cause. They state there are three main actors that bear responsibility for the AI system's actions: users, developers/designers, and authorities. At the end of the day, humans are still responsible for how AI systems behave. Accountability and responsibility are both closely related ideas in this context. Here accountability is the ability to justify decisions produced by the model. These concepts have two main aims, which include explanations and accountability for design.
    
    \subsubsection {Privacy}
\hfill\\
This is one of the most important concepts in ethical AI development. UNESCO lists privacy as one of the 10 core principles of AI Ethics \cite{noauthor_ethics_nodate}. They state privacy should be protected and promoted throughout the life-cycle of an AI model. Jobin et al. \cite{jobin_global_2019} describe this concept as not only a fundamental value but also a right to be protected. It is often discussed in the context of data protection and security, which ensures users' personal or sensitive information is safeguarded \cite{jobin_global_2019}. The authors in \cite{werthner_introduction_2024}, firmly believe that AI systems must consistently protect individual's privacy and never store any data, especially those considered sensitive.
\hfill\\
\hfill\\
Jobin et al. \cite{jobin_global_2019} explain how even though there is a general agreement on these principles there can be divergence in their respective interpretations. The way these principles are applied to different areas and implementation methods can also vary.
One thing to note is that although there are detailed ethical guidelines provided by private or public organizations, the main challenge lies in incorporating them into the development and deployment process of the system. This challenge can arise because of various reasons including vagueness of the guidelines, accuracy trade-offs, technical debt, feasibility and system complexity. Despite the challenges, designers should still aim to produce models that consider ethical guidelines.

.

\section{Challenges and Limitations}

Even though the approaches discussed earlier for mitigating bias show promising results, they have their own challenges and limitations. These approaches and techniques can prove to be unethical in settings where the accuracy of the model is crucial. Integrating fairness constraints to mitigate model bias can cause performance loss and vice versa \cite{zliobaite2015relation,hardt2016,haas19}. However, it is important to realize that how the model is evaluated is also a big concern. Model evaluation techniques can conceal or create new underlying biases. Establishing a balance between fairness and accuracy in such a model is a multi-faceted challenge. There is no one solution that is applicable to all cases. Thus, the ideal balance between these two concepts depends on their distinctive setting and application. Another challenge arises because there is no one universally accepted definition of fairness that can be utilized for all different AI models. The authors \cite{barocas_big_2016} state that maintaining group fairness may result in unequal treatment towards individuals while emphasizing individual fairness may not handle systematic bias at the group level. It can also be challenging to recognize which definition of fairness is relevant for a particular context. Moreover, these definitions are updated and or changed over time and can pose a problem in the interpretation and fulfillment of each of them. Developers can overlook some intricate definitions because of their complexity. Additionally they tend to stick to satisfying demographic parity because of its simplicity. 
\hfill\\
The approaches and techniques to mitigate bias mostly rely on statistical methods and might fail to encapsulate the nuances of the intricate details of human behaviors and decision-making. One of the ways to develop fair models is by incorporating inclusive datasets that represent the diversity of the world fairly. However, having a truly inclusive dataset is a challenge on its own. Some mitigating techniques employ human intervention. Although it is a promising solution, addressing bias this way has many limitations. These include the fact that human interventions can be subjective and can be prone to bias. Since humans are given the power to define bias and report it, this can be a conflict of interest. Also, it may lead to a lack of consistency since different individuals react in different ways, which can also be an issue. The authors in \cite{calegari23} state that although human judgment can help mitigate bias, the process of rectifying bias from models is easier than addressing human prejudice. A pressing issue, pointed out by the authors in \cite{lin_engineering_2021}, is how simply treating individuals equally is not enough to satisfy the notion of fairness, and the concept of equity should also be considered. Only considering the three stages (pre, in and post-processing) when developing mitigation techniques can be limiting. Microsoft has formulated a more modern approach with nine stages that provide a more sturdy process to evaluate fairness \cite{amershi2019software}. Regardless of the challenges and limitations, the development and deployment of fair and unbiased automated systems is a continuous process. Future work can address these challenges and limitations whilst continuing on new innovations that grasp the subtleties of fairness and equity in AI.




\section{Conclusion }
In this survey our focus lies in presenting the main ideas and research conducted to make AI models more fair and ethical. Even with having this specific focus, the amount of pertinent studies is vast, and the aim of this paper is not to provide an overview of all these efforts but rather to address key takeaways from several recurring themes and areas that offer valuable insights for our readers. We have established the significance of this survey by mentioning how it differs from other related surveys. To promote a better understanding of the concept of fairness and bias in the context of AI, we discuss the numerous definitions of fairness and types of bias. 
An intriguing observation is that the majority of the biases are data-driven which can be mitigated using techniques like sampling, causal reasoning and re-weighting. Moreover, some of these biases can arise in multiple stages of the ML pipeline which puts emphasis on the need of a multi-pronged approach when trying to mitigate them. A common theme that we picked up was how researchers like to combine different mitigation strategies to gain a more holistic approach when addressing bias in models. Out of all the different mitigation strategies discussed, adversarial learning and regularization stand out for their wide-spread adoption. Additionally we also cover less common approaches with potential, like RL (Reinforcement Learning) and calibration.
\hfill\\
 Next we dissect real-world cases of models that manifested discriminatory practices within different sectors. Across these sectors, a persistent pattern of bias against individuals from certain races (especially Black people) is observed. This finding is a jarring reminder of the need to investigate the causes of bias in AI models. Additionally, we talk about the ramifications biased models have on users. Exploring this domain is crucial since it helps researchers develop models that are not only fair but also equitable. We also delve into some ethical guidelines, recognizing them as somewhat of a roadmap for fair and responsible AI development. These aid developers, policy makers and researchers to work on models that are fair, transparent, accountable and explainable. A challenge when following these guidelines arises because of the vagueness and subject to interpretation nature of some of these guidelines. More research can be done in addressing the ambiguity of these guidelines. Lastly, we mentioned the challenges and limitations of the existing literature to motivate prospective advancements in the field of fair AI models.
This survey emphasizes on the big picture by covering the why, how and what of bias and fairness in AI models. Future research can explore certain areas that fell outside of this paper's scope, which include (but not limited to), the standardized methods for testing for bias in models, less prominent mitigation strategies and methods to make models interpretable and explainable.
 To bring it all together, developing and deploying fair automated models is crucial, especially in mitigating biases that can affect life-altering decisions. Due to the widespread utilization of AI, which expands to all aspects of our lives, demands for fair and ethical automated choices are bound to happen. Thus, researchers should aim to create fair models and try to alleviate bias from them.

\begin{acks}
This material is based upon work supported by the Air Force Office of Scientific Research under award number FA2386-23-1-4003.
\end{acks}

\bibliographystyle{ACM-Reference-Format}
\bibliography{myBib.bib}

\appendix
\section{Appendices}


\subsection{Fair AI Solutions}

Designing solutions for biased models, cater for several concepts like responsible AI, explainability, transparency, accountability and interpretability \cite{cheng2021socially}. Regardless of the distinct ethical issues they tackle, they all share a unified aim of developing "Fair AI" \cite{richardson2021framework}. As mentioned before, automated tools are being utilized increasingly in making critical decisions surrounding individuals and as such raises concerns about any bias present. The sensitivity of this issue has propelled researchers to come up with solutions for it. Currently, there are several software toolkits available for achieving "Fair AI". We are going to talk about the most popular ones.
\subsubsection{}{IBM's AI Fairness 360 \cite{Bellamy19}}
\hfill\\
AIF360 is an open-source python toolkit that addresses concerns surrounding bias in AI models. The aim of the developers include identifying and quantifying potential bias in datasets used to train models, examining the sources of bias and employing various techniques to mitigate the bias. They utilize a thorough set of fairness metrics including (but not limited to) statistical parity and equalized odds. They also provide extensive explanations for these fairness metrics. Their bias mitigation algorithms include techniques like adversarial de-biasing, group fairness optimization and more. The framework itself is flexible and lets users integrate their own algorithms. Moreover it provides a user-friendly interface and a testing infrastructure to ensure code is reliable. In general, this tool is a valuable addition to the solution space of fair AI.

\subsubsection{LinkedIn's Fairness Toolkit LiFT \cite{Vasudevan2020LiFTAS}}
\hfill\\
LiFT is developed to deal with the issue of fairness and bias in large ML models. The main idea of this tool is to detect and mitigate bias in models that are used for making critical decisions. One of the main advantages of this toolkit is that it leverages Apache Spark to effectively manage large datasets. Like AIF360, LiFT also caters for various fairness metrics including (but not limited to) statistical parity and equalized odds. It investigates training data and detects bias based on sensitive attributes. It utilizes various mitigation strategies including post-processing to reduce bias and promote fairness. It provides flexibility with integrating this into the different stages of the ML pipeline. Additionally the APIs for LiFT, are user-friendly and easy to implement. All in all, this toolkit is a great option that ensures fairness in applications, is scalable and provides accountability and interpretability. 

\subsubsection{Google's What-If Toolkit WIT \cite{wexler19}}
\hfill\\
This tool strives to address the two main challenges of evaluating an ML model: the difficulty in interpreting how the model produces an outcome and the lack of diverse testing (using inclusive dataset and hypothetical scenarios). This tool lets the users modify inputs and investigate how that changes the outcome, essentially giving the opportunity to explore hypothetical scenarios. Users can also explore the model's behavior by utilizing feature importance analysis and visualizations. A perk of this tool is that it requires minimal coding so it proves to be helpful not only to people with technical background but also to those without. This tool enables users to acquire valuable insight about the model being used, and in turn promotes transparency, reliability and interpretability.
\hfill\\
\hfill\\
The authors in \cite{richardson2021framework} extensively discusses about fairness solutions in practice for AI bias, . This paper can provide more in-dept knowledge in this domain.
\subsection{Paper selection Process for Graph}\label{psp}
The first step to this process was to use the query string:
\hfill\\
\textit{fair OR fairness OR bias AND ( ai OR artificial AND intelligence OR machine AND learning OR ml ) AND PUBYEAR > 2016 AND PUBYEAR < 2024 AND ( LIMIT-TO ( LANGUAGE , "English" ) ) AND ( LIMIT-TO ( DOCTYPE , "ar" ) OR LIMIT-TO ( DOCTYPE , "cp" ) OR LIMIT-TO ( DOCTYPE , "ch" ) OR LIMIT-TO ( DOCTYPE , "bk" ) ) AND ( LIMIT-TO ( EXACTKEYWORD , "Machine Learning" ) OR LIMIT-TO ( EXACTKEYWORD , "Artificial Intelligence" ) ) }
\hfill\\
in the advanced search section of scopus. The papers were then exported as csv and categorized using the GPT-extension (in Google Sheets) to ensure they are indeed related to the domain of fairness, bias and/or XAI. Then the number of papers were plotted against the years.









\end{document}